\documentclass[runningheads]{llncs}

\usepackage{eccv}

\usepackage{eccvabbrv}
\usepackage{graphicx}
\usepackage{booktabs}
\usepackage{amsmath}
\usepackage{amssymb}
\usepackage{multirow}
\usepackage{xcolor}
\usepackage{array}
\usepackage{colortbl}
\usepackage{enumitem}
\usepackage{subcaption}
\usepackage{algorithm}
\usepackage{algorithmic}
\usepackage{tabularx}
\usepackage{listings}
\lstdefinestyle{prompt}{
  basicstyle=\ttfamily\scriptsize,
  breaklines=true,
  breakatwhitespace=false,
  frame=single,
  framerule=0.5pt,
  rulecolor=\color{gray},
  backgroundcolor=\color{gray!10},
  xleftmargin=5pt,
  xrightmargin=5pt,
  aboveskip=5pt,
  belowskip=5pt,
}
\usepackage[accsupp]{axessibility}

\usepackage[table]{xcolor}
\definecolor{frameworkgray}{gray}{0.92}
\definecolor{sectiongray}{gray}{0.96}
\newcolumntype{L}{>{\raggedright\arraybackslash}X}
\newcommand{\method}[1]{\hspace{1.2em}#1}

\usepackage{hyperref}

\usepackage{orcidlink}

\newcommand{\meissa}{\textsc{Meissa}}

\renewcommand{\eg}{\mbox{e.g.,\ }}
\renewcommand{\etal}{\mbox{et al.}}

\definecolor{lightgray}{gray}{0.92}
\definecolor{lightblue}{RGB}{230,240,255}
\definecolor{lightgreen}{RGB}{230,255,230}

\begin{document}

\title{\meissa{}: Multi-modal Medical Agentic Intelligence}

\titlerunning{\meissa{}: Multi-modal Medical Agentic Intelligence}


\author{Yixiong Chen\inst{1} \and Xinyi Bai\inst{2} \and Yue Pan\inst{1} \and Zongwei Zhou\inst{1} \and Alan Yuille\inst{1}}
\authorrunning{Chen et al.}
\institute{
Johns Hopkins University, Baltimore, MD, USA\\
\email{ychen646@jh.edu}
\and
Cornell University, Ithaca, NY, USA
}

\maketitle

\begin{abstract}
Multi-modal large language models (MM-LLMs) have shown strong performance in medical image understanding and clinical reasoning. Recent medical agent systems extend them with tool use and multi-agent collaboration, enabling complex decision-making beyond single-pass inference.
However, these systems rely almost entirely on proprietary frontier models (\eg GPT), whose API-based deployment incurs high cost, high latency, and privacy risks that conflict with on-premise clinical requirements.
We present \meissa{}, a lightweight 4B-parameter medical MM-LLM that brings full agentic capability offline. Instead of imitating static answers, \meissa{} learns both \emph{when} to engage external interaction (strategy selection) and \emph{how} to execute multi-step interaction (strategy execution) by distilling structured trajectories from frontier agent systems.
To be specific, we propose:
(1) \textbf{Unified trajectory modeling}: trajectories (reasoning and action traces) are represented within a single state–action–observation formalism, allowing one model to generalize across heterogeneous medical environments.
(2) \textbf{Three-tier stratified supervision}: the model’s own errors trigger progressive escalation from direct reasoning to tool-augmented and multi-agent interaction, explicitly learning difficulty-aware strategy selection.
(3) \textbf{Prospective–retrospective supervision}: pairing exploratory forward traces with hindsight-rationalized execution traces enables stable learning of effective interaction policies.
Trained on ${\sim}$40K curated trajectories, \meissa{} matches or exceeds proprietary frontier agents in 10 of 16 evaluation settings across 13 medical benchmarks spanning radiology, pathology, and clinical reasoning. Using over 25× fewer parameters than typical frontier models like Gemini-3, \meissa{} operates fully offline with ${\sim}22\times$ lower end-to-end latency compared to API-based deployment.
Data, models, and environments are released at https://github.com/Schuture/Meissa.

\keywords{Agentic Behavior Distillation \and Multi-modal Medical Agent \and Tool-Augmented Reasoning \and Visual Question Answering}
\end{abstract}

\section{Introduction}
\label{sec:intro}

Multi-modal large language models (MM-LLMs) can combine medical images with free-form clinical text, enabling unified visual understanding and clinical reasoning~\cite{yao2022react,schick2023toolformer,shen2023hugginggpt}. When extended with \emph{agentic capabilities}, the ability to take external actions such as invoking tools, requesting intermediate evidence, and coordinating specialist roles, these models can solve clinical tasks that are difficult for single-pass prediction systems~\cite{kim2024mdagents,fallahpour2025medrax,schmidgall2024agentclinic}. However, today’s strongest medical agents are usually powered by proprietary frontier models (\eg GPT~\cite{hurst2024gpt}, Gemini~\cite{team2023gemini}, Claude~\cite{anthropic2024claude3}) through cloud APIs. This deployment pattern is often incompatible with real clinical constraints: patient data must remain on-premise, repeated API calls increase cost, and multi-step interaction adds latency that can disrupt clinical workflows.

This paper is motivated by a simple tension: frontier agents are capable, but hard to deploy. We ask:
\emph{Can we transfer the action-taking behavior of frontier medical agents into a model that can run fully offline?}
Here, \emph{agentic behavior} refers to a model’s \emph{interaction policy}: deciding whether to answer directly or to take external actions, and if it acts, choosing a sequence of actions based on observed outcomes. This differs from prior works distilling single-pass \emph{reasoning ability}~\cite{wei2022chain,gunasekar2023textbooks,zhou2023lima,saab2024capabilities} from larger models to smaller models, and from those learning \emph{tool-use} in a single fixed pipeline~\cite{zheng2025deepeyes,hong2025deepeyesv2,jiang2025ophiuchus}, because practical medical settings involve multiple interaction modes (direct reasoning, tool chains, visual feedback loops, role-based consultation, and multi-turn clinical questioning).

\textbf{The core challenge is the lack of structured supervision for agentic behavior.}
Although frontier models exhibit strong agentic behaviors, there is currently no publicly available training corpus designed to teach models how to behave as medical agents. Existing benchmarks provide final answers but do not specify whether external interaction was necessary or how actions should evolve based on intermediate observations. On the other hand, existing agent systems generate agentic trajectories (reasoning and action traces) within isolated environments, each with its own format and protocol, making it difficult to construct unified supervision across heterogeneous interaction modes.

To address this gap, we design a data synthesis pipeline that systematically constructs supervision for agentic behavior along three dimensions.
\textbf{First}, we introduce a unified \emph{state--action--observation} representation that models multiple interaction modes within a single formalism. Therefore, we avoid environment-specific formats and enable cross-environment learning within one model.
\textbf{Second}, to teach \emph{when} to act, we propose a \emph{three-tier stratification strategy} driven by the model’s own errors. Queries that the model already solves become direct-reasoning trajectories (tier 1); queries solvable by a stronger frontier model without tools become enhanced reasoning trajectories (tier 2); and only the hardest residual samples trigger full agent environments (tier 3). This stratified supervision implicitly encodes the decision boundary between direct inference and external interaction.
\textbf{Third}, to teach \emph{how} to act within environments, we construct complementary \emph{prospective} and \emph{retrospective} trajectories. Prospective traces are recorded during forward interaction and capture exploratory decision-making under real observations. Retrospective traces preserve the same action sequence but provide clean hindsight explanations. Together, they provide stable execution supervision across heterogeneous medical agent environments.

Based on this pipeline, we train \meissa{}\footnote{Named after $\lambda$~Orionis (Meissa), the central anchor of the Orion constellation, our model functions as a unified core that coordinates heterogeneous tools and sub-agents in various medical agent environments.}, a 4B-parameter multi-modal medical agent initialized from Qwen3-VL-4B model and distilled from approximately 40K frontier multimodal agent's (Gemini-3-flash) trajectories.
Training requires only ${\sim}$12 hours on 8$\times$A6000 GPUs. Despite its lightweight size, \meissa{} is competitive with frontier models using over 25× fewer parameters across 13 medical benchmarks (\cref{sec:main_results}), matches or exceeds them in 10 of 16 evaluation settings, achieves near-oracle strategy selection, and reduces end-to-end latency by approximately 22$\times$ (\cref{sec:routing}) compared to cloud-based agent deployment.

In summary, our work makes three contributions:
(1) we introduce a unified trajectory representation and targeted synthesis pipeline for constructing supervision of medical agentic behavior across heterogeneous environments;
(2) we show \meissa{}, a lightweight medical agentic model, can make complex agentic behaviors available for offline deployment with low cost and inference latency;
(3) we provide extensive empirical evidence that \meissa{} learns meaningful policy and matches closed-source models on both in-distribution and out-of-distribution (OOD) benchmarks.

\section{Related Work}
\label{sec:related}

\subsection{Medical Multi-modal Large Language Models}
Medical MM-LLMs have evolved from specialized architectures~\cite{li2023llava,luo2023biomedgpt,chen2024chexagent,lin2023pmc,wu2024pmc,luo2022biogpt} to adaptations of general-purpose foundations~\cite{saab2024capabilities,yang2024advancing,tu2024towards}. LLaVA-Med~\cite{li2023llava} demonstrated biomedical VQA through instruction-tuned visual chat~\cite{liu2023visual}, CheXagent~\cite{chen2024chexagent} specialized in chest X-ray interpretation, and HuatuoGPT-Vision~\cite{chen2024towards} injected medical visual knowledge at scale. Med-PaLM~\cite{singhal2023large,singhal2025toward}, Med-Gemini~\cite{tu2024towards}, and RadFM~\cite{wu2025towards} further pushed performance through model scaling and proprietary data. However, these models operate as single-pass reasoners without the ability to invoke tools or adaptively route queries based on difficulty.

\subsection{Tool-Augmented Reasoning and Its Medical Applications}
ReAct~\cite{yao2022react} established reasoning-action interleaving; FireAct~\cite{chen2023fireact} and AgentTuning~\cite{zeng2024agenttuning} showed that fine-tuning on agent trajectories generalizes agentic capabilities across tasks. In the visual domain, DeepEyes~\cite{zheng2025deepeyes} incentivized ``thinking with images'' via RL, and Ophiuchus~\cite{jiang2025ophiuchus} adapted this for medical imaging through cold-start SFT plus agentic RL. In the medical domain, MedRAX~\cite{fallahpour2025medrax} integrated specialized tools for chest X-ray reasoning. Closely related, STeP~\cite{chen2025training} and STAR~\cite{wu2026internalizing} train student models from teacher trajectories via trajectory-aligned SFT, and MedVistaGym~\cite{lu2026medvistagym} frames tool-augmented medical reasoning as an RL problem. Our work differs in distilling agentic behavior \emph{across four heterogeneous environments} into a single model, and in showing that SFT alone can match RL pipelines at substantially lower compute.

\subsection{Multi-Agent Systems for Medicine}
MDAgents~\cite{kim2024mdagents} dispatches queries to solo experts or multi-agent debate panels based on difficulty; MedAgentSim~\cite{almansoori2025medagentsim} and AgentClinic~\cite{schmidgall2024agentclinic} created simulated clinical environments for multi-turn diagnosis. Park \etal~\cite{park2023generative} demonstrated emergent behaviors from generative agent simulations, Du \etal~\cite{du2024improving} showed debate improves factual accuracy, and Society of Mind~\cite{zhuge2025mindstorms} explored hierarchical organization. All require frontier LLM backends at inference. We instead distill multi-agent interaction policies into a single model.

\subsection{Learned Routing, Selective Prediction, and Early Exit}
A model that can invoke external tools must also learn \emph{when not to}. RouteLLM~\cite{ongroutellm} and cascade routing~\cite{dekoninckunified} train lightweight routers to select between a cheap and an expensive LLM; confidence-token methods~\cite{chuang2025learning} embed routing inside the model itself. Selective classification~\cite{geifman2017selective} provides theoretical grounding for risk-coverage trade-offs. Our work differs in learning routing \emph{implicitly} via stratified trajectory supervision. The model's generated reasoning trajectory determines interaction depth without an explicit router, and in treating routing as a continuous choice over depth ($T{=}0$ to $T{>}10$) across four heterogeneous environments, a setting not addressed by existing frameworks.

\section{Method}
\label{sec:method}

\begin{figure*}[t]
  \centering
  \includegraphics[
    width=\linewidth,
    trim=1cm 4cm 1cm 4cm,
    clip
]{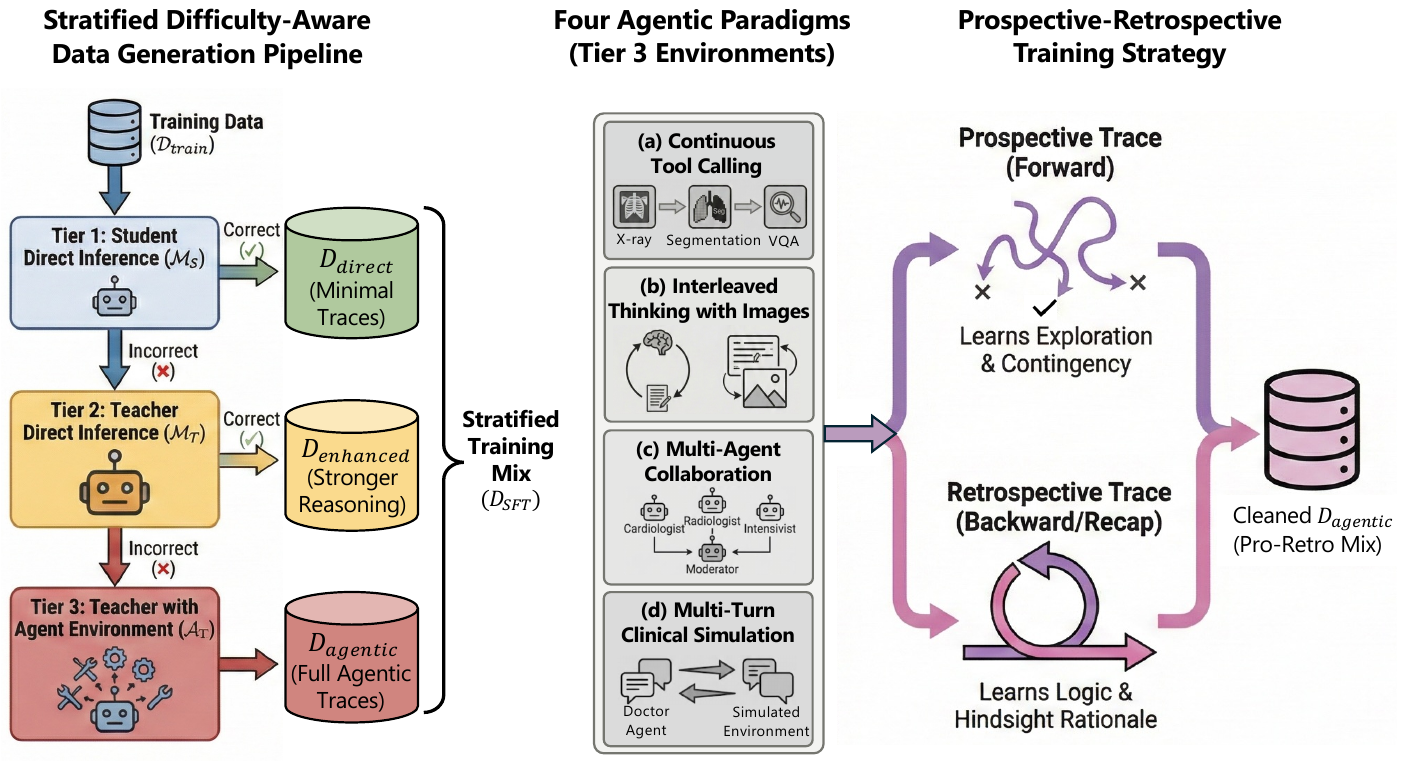}
  \caption{\textbf{Overview of \meissa{}: Trajectory-based agentic behavior distillation.}
  \textbf{Left:} Stratified trajectory supervision uses the model's own errors to progressively escalate interaction depth, teaching strategy selection.
  \textbf{Center:} Four agent environments serve as diverse trajectory sources.
  \textbf{Right:} Prospective-retrospective supervision teaches both exploration and optimal execution policies.}
  \label{fig:overview}
  \vspace{-0.3cm}
\end{figure*}

We frame the training of \meissa{} as \emph{agentic behavior distillation}: learning a policy over \emph{interaction depth}, which is the number of external actions an agent takes before producing an answer. This single variable unifies direct reasoning ($depth=0$), single-tool queries ($depth=1$), multi-step tool chains, multi-agent debate, and clinical simulation into a continuous spectrum. We first define a trajectory formalism that captures this spectrum (\cref{sec:behavior_modeling}), then describe how stratified trajectories teach strategy selection (\cref{sec:hierarchical}), the agent environments that serve as trajectory sources (\cref{sec:paradigms}), and the prospective-retrospective supervision that teaches strategy execution (\cref{sec:proretro}). \Cref{fig:overview} provides an overview.

\subsection{Agentic Behavior Modeling}
\label{sec:behavior_modeling}

We model all agent interactions from direct reasoning to complex multi-agent collaboration within a unified trajectory formalism. An \emph{agentic trajectory} is a sequence:
\begin{equation}
\tau = [(s_0, a_0, o_1), (s_1, a_1, o_2), \ldots, (s_{T-1}, a_{T-1}, o_T)],
\label{eq:trajectory}
\end{equation}
where state $s_t$ is the conversation context up to step $t$, $a_t$ is an action, and $o_t$ is the observation returned by the environment. Concretely, each action $a_t$ is serialized as either a \texttt{<|call|>} turn (a JSON-formatted action) or an \texttt{<|assistant|>} turn for the final answer. \texttt{<|observation|>} turn is returned by tools or sub-agents in JSON format. Inference terminates when the model emits an \texttt{<|assistant|>} turn or reaches $T_{\max}$; full schemas are in the supplementary.

The \emph{interaction depth} $T$ of a trajectory is the key variable that unifies all agent behaviors. Direct reasoning corresponds to $T{=}0$ (a single final-answer). For \meissa{}, we set $T_{\max}=4$ for continuous tool calling; $T_{\max}=6$ for interleaved image-text reasoning; and multi-agent collaboration and clinical simulation produce the deepest trajectories ($T_{\max}=12$). Learning \emph{when} to produce a trajectory of a given depth, and \emph{how} to act at each step, is the core objective of agentic behavior distillation.

We train the policy $\pi_\theta(a_t \mid s_t)$ based on trajectories generated by a frontier model, where next-action prediction reduces to conditional language modeling. This separates the \emph{learning problem} from the \emph{environment details}, enabling a single model to generalize across environments. Critically, behavioral cloning over \emph{ordered} trajectories learns causal decision-making: the model predicts each action conditioned on preceding observations. (Verified on \cref{sec:ablation_robustness}).

\subsection{Behavior Distillation via Stratified Trajectory Supervision}
\label{sec:hierarchical}

The key insight for teaching strategy selection is that not all queries require the same interaction depth. Rather than uniformly generating complex agentic trajectories for all samples, we stratify trajectory supervision using the model's own errors as a curriculum signal~\cite{bengio2009curriculum}:

\noindent\textbf{Tier~1: Direct Reasoning Trajectories.} We first run the student model $\mathcal{M}_S$ (Qwen3-VL-4B~\cite{bai2025qwen3}) directly on the training set $\mathcal{D}_{\text{train}}$. Samples answered correctly yield direct reasoning trajectories ($T{=}0$, no external interaction):
\begin{equation}
\mathcal{D}_{\text{direct}} = \{(x, \tau_{\text{direct}}) \mid (x, y) \in \mathcal{D}_{\text{train}},\; \mathcal{M}_S(x) = y\}.
\end{equation}
These trajectories supervise the model to answer efficiently when its parametric knowledge suffices, learning \emph{when not to act}.

\noindent\textbf{Tier~2: Enhanced Reasoning Trajectories.} The student's incorrect samples are re-processed by a stronger teacher model $\mathcal{M}_T$ (Gemini-3-flash \cite{team2023gemini}), again without external interaction:
\begin{equation}
\mathcal{D}_{\text{enhanced}} = \{(x, \tau_{\text{enhanced}}) \mid (x, y) \in \mathcal{D}_{\text{train}} \setminus \mathcal{D}_{\text{direct}},\; \mathcal{M}_T(x) = y\}.
\end{equation}
These trajectories demonstrate stronger reasoning that the student could not produce, but remain direct reasoning ($T{=}0$).

\noindent\textbf{Tier~3: Agentic Trajectories.} The hardest residual samples are beyond both the student's and teacher's direct reasoning capacity. They are processed by the teacher within agent environments (\cref{sec:paradigms}):
\begin{equation}
\mathcal{D}_{\text{agentic}} = \{(x, \tau_{\text{agentic}}) \mid (x, y) \in \mathcal{D}_{\text{residual}},\; \mathcal{A}_T(x) = y\},
\end{equation}
where $\mathcal{A}_T$ denotes the teacher operating within an agent environment and residual erroneous samples $\mathcal{D}_{\text{residual}} = \mathcal{D}_{\text{train}} \setminus (\mathcal{D}_{\text{direct}} \cup \mathcal{D}_{\text{enhanced}})$. Only these samples produce full multi-step trajectories with external interaction.

The complete data generation procedure, including retry logic, per-tier statistics, and discard handling, is detailed in Algorithm~1 in supplementary material. The final dataset contains ${\sim}$40K trajectories: 8.2K direct (Tier~1), 9.8K enhanced (Tier~2), and 23.9K agentic (Tier~3).

The combined supervision teaches strategy selection as an emergent behavior: simple queries are associated with direct trajectories, complex queries with multi-step ones, naturally inducing a difficulty-aware routing policy. At inference, the model's first generated action implicitly determines the strategy.

\subsection{Agent Environments and Trajectory Sources}
\label{sec:paradigms}

\begin{figure}[t]
  \centering
  \includegraphics[
    width=\linewidth,
    trim=1cm 4cm 1cm 4cm,
    clip
]{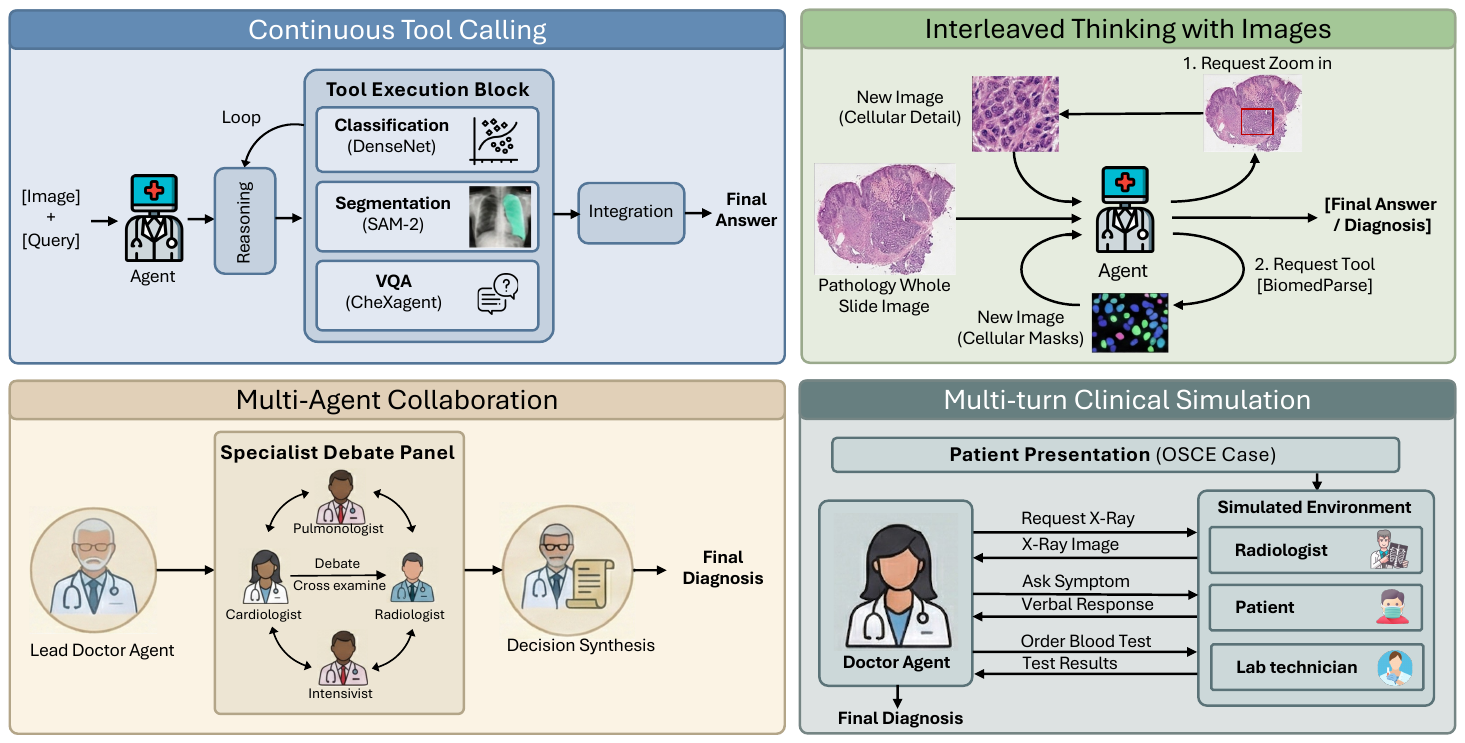}
  \caption{\textbf{Four agent environments as trajectory sources.}
  Each environment produces trajectories with distinct state--action--observation patterns:
  (a)~tool calling trajectories with vision tool chains,
  (b)~interleaved image-text trajectories with visual feedback loops,
  (c)~multi-agent trajectories with expert debate and synthesis,
  (d)~clinical simulation trajectories with multi-turn information gathering.}
  \label{fig:paradigms}
  \vspace{-0.5cm}
\end{figure}

The agentic trajectories in $\mathcal{D}_{\text{agentic}}$ are collected from four agent environments, each contributing trajectories with distinct interaction patterns (\cref{fig:paradigms}). These environments serve as \emph{trajectory sources} rather than architectural components of \meissa{} itself. Their diversity ensures that the learned policy generalizes.

\subsubsection{Visual information in trajectories.} In tool-calling trajectories, the initial state $s_0$ contains the input image as \texttt{<image>} tokens; observations return structured text (classification labels, bounding boxes, reports). In interleaved image-text trajectories, observations additionally contain \emph{new images} like cropped regions or segmentation overlays injected as additional \texttt{<image>} tokens, creating a multi-round visual reasoning chain. Multi-agent environments provide \texttt{<image>} tokens for lead agent and each sub-agent at the beginning of the debate. For clinical simulation, \meissa{} can request either structured examination results or images as \texttt{<image>} tokens as needed in each call.

\subsubsection{Action Space of the Agent Environments.}

\begin{itemize}
    \item \noindent\textbf{Continuous Tool Calling~\cite{fallahpour2025medrax}.}
The action space consists of eight medical imaging tools: $a_t \in \{$\texttt{Classifier}, \texttt{Segmentation}, \texttt{CheXagentVQA}, \texttt{LLaVA-Med}, \texttt{PhraseGrounding}, \texttt{ReportGenerator}, \texttt{DICOMProcessor}, \texttt{ImageVisualizer}$\}$, each serialized as a \texttt{<|call|>} turn with JSON arguments. Observations $o_{t+1}$ return structured text: classification probabilities, bounding box coordinates, segmentation statistics, or free-text reports.
    \item \noindent\textbf{Interleaved Thinking with Images~\cite{zheng2025deepeyes,hong2025deepeyesv2,jiang2025ophiuchus}.}
Actions select from vision manipulation tools: $a_t \in \{\texttt{ZoomIn},\allowbreak \texttt{SAM2},\allowbreak \texttt{BioMedParse},\allowbreak \texttt{OCR},\allowbreak \texttt{Terminate}\}$. Unlike tool calling, observations $o_{t+1}$ contain \emph{new images}, creating multi-round visual reasoning chains. The trajectory terminates when the model emits \texttt{Terminate} with the final answer.
    \item \noindent\textbf{Multi-Agent Collaboration~\cite{kim2024mdagents}.}
Actions are structured role-based queries: $a_t$ consists of a difficulty assessment, expert recruitment, expert analysis, multi-round debate contributions, or moderator synthesis. Observations $o_{t+1}$ aggregate expert responses (analysis or debate argument). Debate protocols, persona prompts, and synthesis rules are in the supplementary.
    \item \noindent\textbf{Clinical Simulation~\cite{schmidgall2024agentclinic,almansoori2025medagentsim}.}
Actions select from clinical information-gathering tools: $a_t \in \{\texttt{RequestPhysicalExam},\allowbreak \texttt{RequestTest},\allowbreak \texttt{Terminate}\}$. The initial state $s_0$ contains the patient presentation (demographics, history, symptoms). Observations $o_{t+1}$ return clinical findings (\eg ``Auscultation: bilateral basal crackles''), lab results (\eg ``Troponin I: 0.4~ng/mL''), or radiology images. The trajectory terminates with \texttt{Terminate} for the final diagnosis.
\end{itemize}

\subsection{Prospective-Retrospective Supervision}
\label{sec:proretro}

Learning \emph{when} to act is addressed by stratified trajectories; here we address learning \emph{how} to act effectively within agent environments. The key insight is that a single correct trajectory can be decomposed into two complementary supervision signals that teach different aspects of execution policy.

\noindent\textbf{Prospective (forward) trajectories} are recorded during real-time inference: the teacher reasons forward, potentially revising strategy upon unexpected observations. Only correct trajectories are retained, teaching an \emph{exploration policy}.

\noindent\textbf{Retrospective (backward) trajectories} are generated \emph{given} each prospective trajectory: a recap agent re-narrates the reasoning while preserving the original action sequence, producing a clean post-hoc narrative that teaches an \emph{optimal execution policy}. The combined training set is:
\begin{equation}
\mathcal{D}_{\text{SFT}} = \mathcal{D}_{\text{direct}} \cup \mathcal{D}_{\text{enhanced}} \cup \{\tau^{\text{pro}}(x),\, \tau^{\text{retro}}(x, y) \mid (x,y) \in \mathcal{D}_{\text{agentic}}\}.
\end{equation}
During auditing, retrospective recaps that are over-simplified or over-complicated are discarded; approximately 8\% of generated recaps are filtered. We did not observe significant gains from depth- or difficulty-based weighting in preliminary experiments. A detailed comparison of the two supervision modes and trajectory validation details (structural and behavioral filtering) are in the supplementary.

\section{Experiments}
\label{sec:experiments}

\subsection{Experimental Setup}
\label{sec:setup}

We evaluate \meissa{} across 13 benchmarks spanning four medical domains (\cref{tab:benchmarks}), where five benchmarks (ChestAgentBench, OmniMed, MedXpertQA, NEJM, and NEJM Ext.) are OOD. Training trajectories are generated with the training sets of the datasets, $n$-gram de-contamination ($n{=}8$) confirms zero overlap with testing data. Decontamination operates over question text, and image-level leakage is prevented by using entirely disjoint DICOM/hashing for benchmarks. Dataset details are in the supplementary.

\begin{table}[t]
  \caption{\textbf{Evaluation benchmarks} (13 total across 4 agent environments). $\dagger$~=~strict OOD: excluded from training and verified by $n$-gram de-contamination ($n{=}8$).}
  \label{tab:benchmarks}
  \centering
  \small
  \begin{tabular}{@{}llllc@{}}
    \toprule
    Benchmark & Domain & Modality & Environment & ID/OOD \\
    \midrule
    MIMIC-CXR-VQA~\cite{aasmimic} & Radiology & CXR + text & Tool Calling + Col. & ID \\
    ChestAgentBench~\cite{fallahpour2025medrax} & Radiology & Multi-CXR + text & Tool Calling & OOD$\dagger$ \\
    PathVQA~\cite{he2020pathvqa} & Pathology & Path img + text & Think img + Col. & ID \\
    SLAKE~\cite{liu2021slake} & Mixed & Med img + text & Think img & ID \\
    VQA-RAD~\cite{lau2018dataset} & Radiology & Rad img + text & Think img & ID \\
    OmniMed~\cite{hu2024omnimedvqa} & Mixed & Med img + text & Think img & OOD$\dagger$ \\
    MedXpertQA~\cite{zuo2025medxpertqa} & Clinical & Med img + text & Think img & OOD$\dagger$ \\
    MedQA~\cite{jin2021disease} & Clinical & Text only & Collaboration + Sim. & ID \\
    PubMedQA~\cite{jin2019pubmedqa} & Biomedical & Text + context & Collaboration & ID \\
    NEJM~\cite{schmidgall2024agentclinic} & Clinical & Text + img & Simulation & OOD$\dagger$ \\
    NEJM Ext.~\cite{schmidgall2024agentclinic} & Clinical & Text + img & Simulation & OOD$\dagger$ \\
    MIMIC-IV~\cite{johnson2023mimic} & Clinical & EHR + CXR & Simulation & ID \\
    MedQA Ext.~\cite{almansoori2025medagentsim} & Clinical & Text only & Simulation & ID \\
    \bottomrule
  \end{tabular}
\end{table}

\noindent\textbf{Baselines.} (1)~Closed-source frontier models: GPT-4o~\cite{hurst2024gpt}, GPT-5~\cite{openai2025gpt5}, Gemini-3-flash~\cite{team2023gemini}; (2)~Agentic frameworks: MedRAX~\cite{fallahpour2025medrax}, Ophiuchus~\cite{jiang2025ophiuchus}, MDAgents~\cite{kim2024mdagents}, and AgentClinic~\cite{schmidgall2024agentclinic}; (3)~Base model: Qwen3-VL-4B-Instruct~\cite{bai2025qwen3}; (4)~Answer-only SFT: Qwen3-VL-4B fine-tuned on the same ${\sim}$40K queries with only teacher final answers, controlling for data exposure.

\noindent\textbf{Implementation.} All experiments use a node with 8 NVIDIA A6000 GPUs (48\,GB). Training: LLaMA-Factory~\cite{zheng2024llamafactory} with LoRA~\cite{hu2022lora} rank 32, learning rate $5\times10^{-5}$, ${\sim}$12 GPU-hours, ${\sim}$239M tokens for 3 epochs. Inference: vLLM~\cite{kwon2023efficient}. All \meissa{} inference results are mean over 3 seeds ($\sigma$ in supplementary). All tasks are formulated as QA or VQA; the evaluation metric is accuracy throughout: exact-match accuracy for close-ended questions and soft-match accuracy with medical synonym normalization~\cite{aasmimic} for open-ended VQA.

\subsection{\meissa{} Is Competitive with Frontier Models}
\label{sec:main_results}

\Cref{tab:main_results} presents results across all 13 benchmarks under four agent evaluation frameworks. \meissa{} is ranked 1st or 2nd on 10 out of 16 evaluation settings, matching frontier models. We summarize more detailed findings below:

\begin{table*}[!t]
\caption{\textbf{Comparison under four agent evaluation frameworks.}
\textbf{Bold}: best per column; \underline{underline}: second-best per column. Parenthetical superscripts on \meissa{} rows denote absolute improvement ($\Delta$) over the untrained Qwen3-VL-4B baseline within the same agent framework. $\dagger$~=~strict OOD benchmark. All Qwen3-VL and Gemini-3-flash results are from our own evaluation, all other results are from the original papers.}
\label{tab:main_results}
\centering
\scriptsize
\renewcommand{\arraystretch}{1.06}

{\setlength{\tabcolsep}{6pt}
\begin{tabularx}{\textwidth}{@{}Lcc@{}}
\toprule
\rowcolor{frameworkgray}
\multicolumn{3}{@{}l@{}}{\textbf{Framework I: Continuous Tool Calling}} \\
\midrule
& \textbf{ChestAgentBench$\dagger$} & \textbf{MIMIC-CXR-VQA} \\
\rowcolor{sectiongray}
\multicolumn{3}{@{}l@{}}{Direct Inference} \\
\method{GPT-4o} & 56.4 & 40.0 \\
\method{Gemini-3-flash} & \textbf{76.2} & 43.6 \\
\method{Qwen3-VL-4B (SFT)} & 53.3 & 49.7 \\
\addlinespace[1pt]
\rowcolor{sectiongray}
\multicolumn{3}{@{}l@{}}{Agent-based (MedRAX)} \\
\method{GPT-4o} & 63.1 & 55.6 \\
\method{Gemini-3-flash} & \underline{72.7} & \underline{65.0} \\
\method{Qwen3-VL-4B} & 46.6 & 51.4 \\
\rowcolor{lightblue}
\method{\textbf{\meissa{} (Ours)}} & 62.8$^{(+16.2)}$ & \textbf{65.2}$^{(+13.8)}$ \\
\bottomrule
\end{tabularx}
}

\vspace{-2.5pt}

{\setlength{\tabcolsep}{3.2pt}
\begin{tabularx}{\textwidth}{@{}Lccccc@{}}
\toprule
\rowcolor{frameworkgray}
\multicolumn{6}{@{}l@{}}{\textbf{Framework II: Interleaved Thinking with Images}} \\
\midrule
& \textbf{PathVQA} & \textbf{SLAKE} & \textbf{VQA-RAD} & \textbf{OmniMed$\dagger$} & \textbf{MedXpertQA$\dagger$} \\
\rowcolor{sectiongray}
\multicolumn{6}{@{}l@{}}{Direct Inference} \\
\method{GPT-5} & 60.0 & 73.2 & 64.5 & 75.4 & 40.4 \\
\method{Gemini-3-flash} & 64.3 & 77.7 & 58.8 & 78.0 & \underline{54.4} \\
\method{Qwen3-VL-4B (SFT)} & 73.0 & 77.1 & 69.1 & 67.6 & 33.3 \\
\addlinespace[1pt]
\rowcolor{sectiongray}
\multicolumn{6}{@{}l@{}}{Agent-based (Ophiuchus)} \\
\method{o3} & 67.5 & 75.3 & 66.0 & 73.7 & 44.1 \\
\method{Gemini-2.5-pro} & 67.1 & 72.7 & 63.8 & 76.9 & 46.6 \\
\method{Gemini-3-flash} & \underline{74.3} & 73.9 & 52.0 & 77.4 & \textbf{69.2} \\
\method{Ophiuchus-7B} & \underline{74.3} & \textbf{83.9} & \textbf{73.6} & \underline{78.6} & 39.3 \\
\method{Qwen3-VL-4B} & 65.3 & 55.6 & 51.8 & 38.1 & 23.9 \\
\rowcolor{lightblue}
\method{\textbf{\meissa{} (Ours)}} & \textbf{78.2}$^{(+12.9)}$ & \underline{82.0}$^{(+26.4)}$ & \underline{70.1}$^{(+18.3)}$ & \textbf{82.8}$^{(+44.7)}$ & 36.0$^{(+12.1)}$ \\
\bottomrule
\end{tabularx}
}

\vspace{-2.5pt}

{\setlength{\tabcolsep}{4.2pt}
\begin{tabularx}{\textwidth}{@{}Lcccc@{}}
\toprule
\rowcolor{frameworkgray}
\multicolumn{5}{@{}l@{}}{\textbf{Framework III: Multi-Agent Collaboration}} \\
\midrule
& \textbf{MedQA} & \textbf{PubMedQA} & \textbf{PathVQA} & \textbf{MIMIC-CXR-VQA} \\
\rowcolor{sectiongray}
\multicolumn{5}{@{}l@{}}{Direct Inference} \\
\method{GPT-4V} & 75.0 & 61.5 & 57.9 & 40.0 \\
\method{Gemini-3-flash} & \underline{75.5} & 66.7 & 64.3 & 43.6 \\
\method{Qwen3-VL-4B (SFT)} & 51.8 & \underline{76.8} & \underline{66.4} & 55.2 \\
\addlinespace[1pt]
\rowcolor{sectiongray}
\multicolumn{5}{@{}l@{}}{Agent-based (MDAgents)} \\
\method{GPT-4V} & \textbf{88.7} & 75.0 & 65.3 & 55.9 \\
\method{Gemini-3-flash} & \underline{75.5} & 71.9 & 56.3 & \textbf{64.0} \\
\method{Qwen3-VL-4B} & 59.8 & 57.3 & 65.5 & 54.9 \\
\rowcolor{lightblue}
\method{\textbf{\meissa{} (Ours)}} & 57.2$^{(-2.6)}$ & \textbf{77.9}$^{(+20.6)}$ & \textbf{67.9}$^{(+2.4)}$ & \underline{59.4}$^{(+4.5)}$ \\
\bottomrule
\end{tabularx}
}

\vspace{-2.5pt}

{\setlength{\tabcolsep}{3.2pt}
\begin{tabularx}{\textwidth}{@{}Lccccc@{}}
\toprule
\rowcolor{frameworkgray}
\multicolumn{6}{@{}l@{}}{\textbf{Framework IV: Multi-turn Clinical Simulation}} \\
\midrule
& \textbf{NEJM$\dagger$} & \textbf{NEJM Ext.$\dagger$} & \textbf{MedQA} & \textbf{MedQA Ext.} & \textbf{MIMIC-IV} \\
\rowcolor{sectiongray}
\multicolumn{6}{@{}l@{}}{Agent-based (AgentClinic)} \\
\method{GPT-4o} & 26.7 & \underline{25.8} & 52.8 & 52.3 & 34.4 \\
\method{Gemini-3-flash} & \underline{40.0} & \textbf{33.3} & \textbf{97.9} & \textbf{92.3} & \underline{70.6} \\
\method{Llama-3.3-70B} & 20.0 & 24.2 & \underline{54.7} & \underline{53.3} & 36.8 \\
\method{Qwen2.5-72B} & 0.0 & 2.5 & 38.7 & 41.6 & 21.2 \\
\method{Qwen3-VL-4B} & \underline{40.0} & 20.8 & 50.5 & 50.0 & 61.1 \\
\rowcolor{lightblue}
\method{\textbf{\meissa{} (Ours)}} & \textbf{46.7}$^{(+6.7)}$ & 23.3$^{(+2.5)}$ & 49.5$^{(-1.0)}$ & 46.7$^{(-3.3)}$ & \textbf{84.4}$^{(+23.3)}$ \\
\bottomrule
\end{tabularx}
}
\vspace{-0.5cm}
\end{table*}

\noindent\textbf{Trajectory supervision advantage.}
\meissa{} surpasses Qwen3-VL-4B by 17.3 points and the answer-only tuned Qwen3-VL-4B direct inference baseline by 6.1 points on average, showing that trajectory supervision conveys substantially richer learning signal than data exposure alone.

\noindent\textbf{OOD robustness.}
On the five strict OOD benchmarks, \meissa{} achieves strong results, most notably 62.8\% on ChestAgentBench and 35.0\% on NEJM (averaged with NEJM Ext.), matching or exceeding GPT-4o despite being 100$\times$ smaller. Gains on ChestAgentBench are most pronounced on \emph{classification} and \emph{detection} sub-categories, where tool-generated classification/grounding information helps enhance the image understanding beyond the Qwen3-VL model itself.

\noindent\textbf{Data-rich VQA tasks.}
\meissa{} achieves the top result on PathVQA (78.2\%; 12.2K training samples) and MIMIC-CXR-VQA (65.2\%; 11.4K samples), ranking first among all methods in both cases. Notably, the base Qwen3-VL-4B already shows competitive baselines on these tasks (65.3\% and 51.4\%), reflecting that data abundance lowers the learning barrier for lightweight models. The gains attributable to \meissa{} thus stem primarily from learned interaction policy rather than bridging a parametric knowledge gap.

\noindent\textbf{Reference-grounded tasks.}
On PubMedQA (77.9\%, first), \meissa{} outperforms GPT-4V under MDAgents by 2.9 points. PubMedQA supplies a reference abstract alongside each question, shifting the demand from encyclopedic recall to evidence synthesis over provided context. This is precisely the execution skill that prospective-retrospective trajectory supervision optimizes, explaining \meissa{}'s disproportionate gain relative to its performance on knowledge-only variants.

\noindent\textbf{Matching RL-trained models at smaller scale.}
Ophiuchus-7B~\cite{jiang2025ophiuchus} is trained via cold-start SFT followed by agentic RL, requiring substantially more compute than our SFT-only pipeline. Despite having nearly half the parameters, \meissa{} matches or exceeds Ophiuchus-7B on 3 of 5 Framework~II benchmarks (PathVQA: 78.2 vs.\ 74.3; OmniMedVQA: 82.8 vs.\ 78.6) while trailing on SLAKE (82.0 vs.\ 83.9) and VQA-RAD (70.1 vs.\ 73.6). This demonstrates that stratified SFT with prospective-retrospective supervision can match RL pipelines.

\noindent\textbf{Knowledge and reasoning limits.}
\meissa{} has clear limitations on tasks requiring deep parametric knowledge or expert clinical reasoning. On MedQA (Framework III), it reaches only 57.2\% against GPT-4V's 88.7\%. On MedXpertQA, it achieves 36.0\% against Gemini-3-flash's 69.2\%, trailing even Ophiuchus-7B (39.3\%). Mild negative transfer also appears on MedQA Ext.\ (Framework IV, $-3.3$\% relative to the base), confirming that agentic trajectory training competes with parametric knowledge retention in lightweight models.


\begin{table}[t]
  \caption{\textbf{Counterfactual strategy comparison on ChestAgentBench.} Learned routing achieves near-oracle accuracy at ${\sim}3\times$ lower cost than always-agentic inference, while avoiding the accuracy degradation caused by over-invoking tools on easy queries. $\ddagger$~=~Gemini-3-flash API does not report token usage; value is a lower-bound estimate.}
  \label{tab:counterfactual}
  \centering
  \small
  \begin{tabular}{@{}lcccc@{}}
    \toprule
    Strategy & Avg Accuracy & Avg Actions & Avg Tokens & Latency \\
    \midrule
    Gemini-3-flash & 72.7 & 4.03 & ${\geq}860^{\ddagger}$ & 87.2s ($72\times$) \\
    \midrule
    Always direct & 54.2 & 1.0 & 479 & 1.23s ($1\times$) \\
    Always agentic & 45.9 & 3.52 & 2082 & 10.64s ($9\times$) \\
    Oracle (upper bound) & 63.2 & 1.54 & 933 & 3.41s ($3\times$) \\
    \rowcolor{lightblue}
    Learned routing (Ours) & 62.8 & 1.71 & 959 & 4.12s ($3\times$) \\
    \bottomrule
  \end{tabular}
\end{table}

\subsection{Strategy Selection: Learning \emph{When} to Act}
\label{sec:routing}

\noindent\textbf{Counterfactual and oracle comparison.} We compare \meissa{}'s learned routing against fixed strategies and an \emph{oracle router} that selects the best strategy per query (\cref{tab:counterfactual}) according to the samples' testing correctness.

\begin{figure}[!t]
  \centering
  \includegraphics[width=\linewidth]{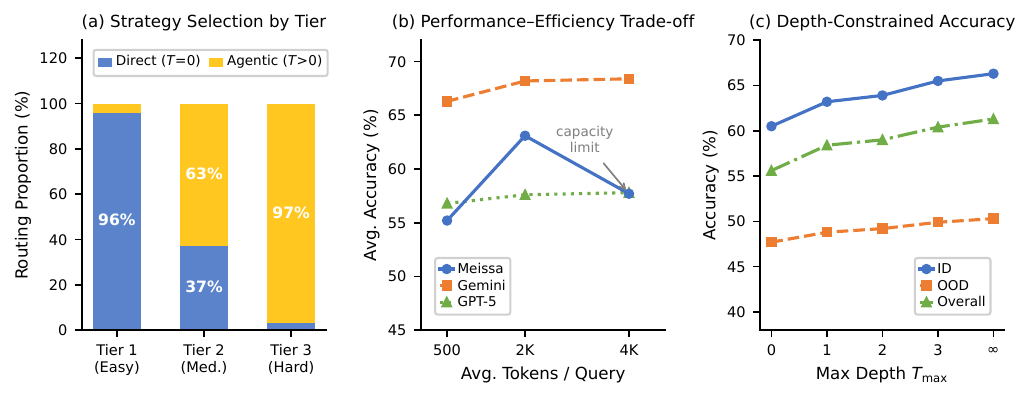}
  \caption{\textbf{Strategy selection analysis.}
  (\emph{Left})~Tier~1 easy queries are answered directly in 96\% of cases, while Tier~3 hard queries trigger agentic interaction 97\% of the time, confirming difficulty-aware routing.
  (\emph{Center})~\meissa{} accuracy peaks near 2{,}000 tokens / query and then drops at 4{,}000 tokens (capacity limit of the 4B model), whereas frontier models scale monotonically. This motivates depth allocation for lightweight models.
  (\emph{Right})~Depth-constrained accuracy ($T_{\max}\in\{0,1,2,3,\infty\}$): accuracy improves consistently with interaction depth but exhibits diminishing returns beyond $T_{\max}{=}3$.}
  \label{fig:routing}
\end{figure}

Learned routing closes the gap to oracle, demonstrating near-optimal strategy selection: 72\% of queries are routed to direct reasoning and 28\% to agentic interaction. Notably, the always-agentic strategy \emph{degrades} average accuracy relative to always-direct, because overlong contexts and occasionally misleading tool outputs hurt easy queries that require no external information. The latency gap is even more striking: Gemini-3-flash averages 87.2s per query due to multiple API calls and remote tool execution, while \meissa{} averages 4.1s, a ${\sim}$22$\times$ speedup. We present the full per-query latency distributions in the supplementary.

\begin{table}[!t]
  \caption{\textbf{Ablation studies.} (a)~Stratified trajectory supervision shows each tier contributes. (b)~Prospective and retrospective supervision are complementary.}
  \label{tab:ablation}
  \centering
  \small
  \begin{tabular}{@{}lccc@{}}
    \toprule
    Configuration & ID Avg & OOD Avg & Overall \\
    \midrule
    \multicolumn{4}{l}{\textit{(a) Stratified trajectory supervision}} \\
    Agentic only (Tier 3) & 61.2 & 46.6 & 56.6 \\
    Direct + Agentic (Tier 1+3) & 59.8 & 46.9 & 55.8 \\
    Enhanced + Agentic (Tier 2+3) & 63.7 & 51.0 & 59.7 \\
    \rowcolor{lightblue}
    Stratified (Tier 1+2+3) & 66.3 & 50.3 & 61.3 \\
    \midrule
    \multicolumn{4}{l}{\textit{(b) Execution policy learning}} \\
    Prospective only & 57.6 & 49.7 & 55.1 \\
    Retrospective only & 66.1 & 47.2 & 60.2 \\
    \rowcolor{lightblue}
    Pro + Retro (Full) & 66.3 & 50.3 & 61.3 \\
    \bottomrule
  \end{tabular}
  \vspace{-0.3cm}
\end{table}

\noindent\textbf{Depth-constrained analysis.} We constrain $T_{\max}$ at inference time (\cref{fig:routing}, right). Overall accuracy rises from 55.6\% ($T_{\max}{=}0$) to 60.4\% ($T_{\max}{=}3$), with only $+0.9$ additional points for $T_{\max}{=}\infty$. It confirms that three interaction rounds capture most of the available gain and deeper interactions exhibit rapidly diminishing returns.

\subsection{Ablation and Robustness: Learning \emph{How} to Act}
\label{sec:ablation}

\subsubsection{Distillation Strategy Ablations.}

\Cref{tab:ablation} shows all three tiers are necessary: Tier~1 preserves parametric knowledge; Tier~2 bridges direct and agentic reasoning (5.5~pt gain over Tier~1+3); Tier~3 provides agentic exemplars. Tier~1+3 is worse than Tier~3 alone due to overfitting on simple samples without Tier~2. For execution policy, prospective-only excels at OOD while retrospective-only excels at ID; their combination is optimal. Teacher sensitivity analysis is in the supplementary.

\subsubsection{Policy Robustness: Causal Decision-Making vs.\ Pattern Imitation.}
\label{sec:ablation_robustness}

\begin{table}[t]
  \caption{\textbf{Policy robustness under perturbation for interleaved thinking with image framework.} $\Delta$ = accuracy change vs.\ unperturbed \meissa{} on agentic-only queries ($T{>}0$). The model relies causally on observation content and visual feedback.}
  \label{tab:ablation_robustness}
  \centering
  \small
  \begin{tabular}{@{}lcccc@{}}
    \toprule
    Perturbation & PathVQA & SLAKE & VQA-RAD & $\Delta$ Avg \\
    \midrule
    \meissa{} (unperturbed) & 75.2 & 80.3 & 67.1 & --- \\
    \midrule
    Paraphrased tool outputs & 76.7 & 76.6 & 66.9 & $-0.8$ \\
    Visual feedback $\to$ blank & 72.0 & 67.2 & 62.9 & $-6.8$ \\
    Visual feedback $\to$ random & 56.6 & 66.5 & 57.8 & $-13.9$ \\
    Tools disabled ($T{=}0$) & 72.1 & 75.7 & 66.3 & $-2.8$ \\
    Altered prompt format & 73.7 & 76.4 & 65.5 & $-2.3$ \\
    \bottomrule
  \end{tabular}
\end{table}

\Cref{tab:ablation_robustness} shows that \meissa{} learns a causal visual policy: paraphrased outputs cause minor drops ($-0.8$), while replacing visual feedback with blank or random images causes much larger drops ($-6.8$ and $-13.9$), confirming that the model \emph{actively perceives} tool-generated observations. Disabling tools drops performance by $-2.8$ on agentic queries; altering prompt format causes only $-2.3$ points, indicating abstraction beyond template memorization.


\begin{figure*}[!t]
  \centering
  \includegraphics[
    width=\linewidth,
    trim=1cm 0cm 1cm 0cm,
    clip
]{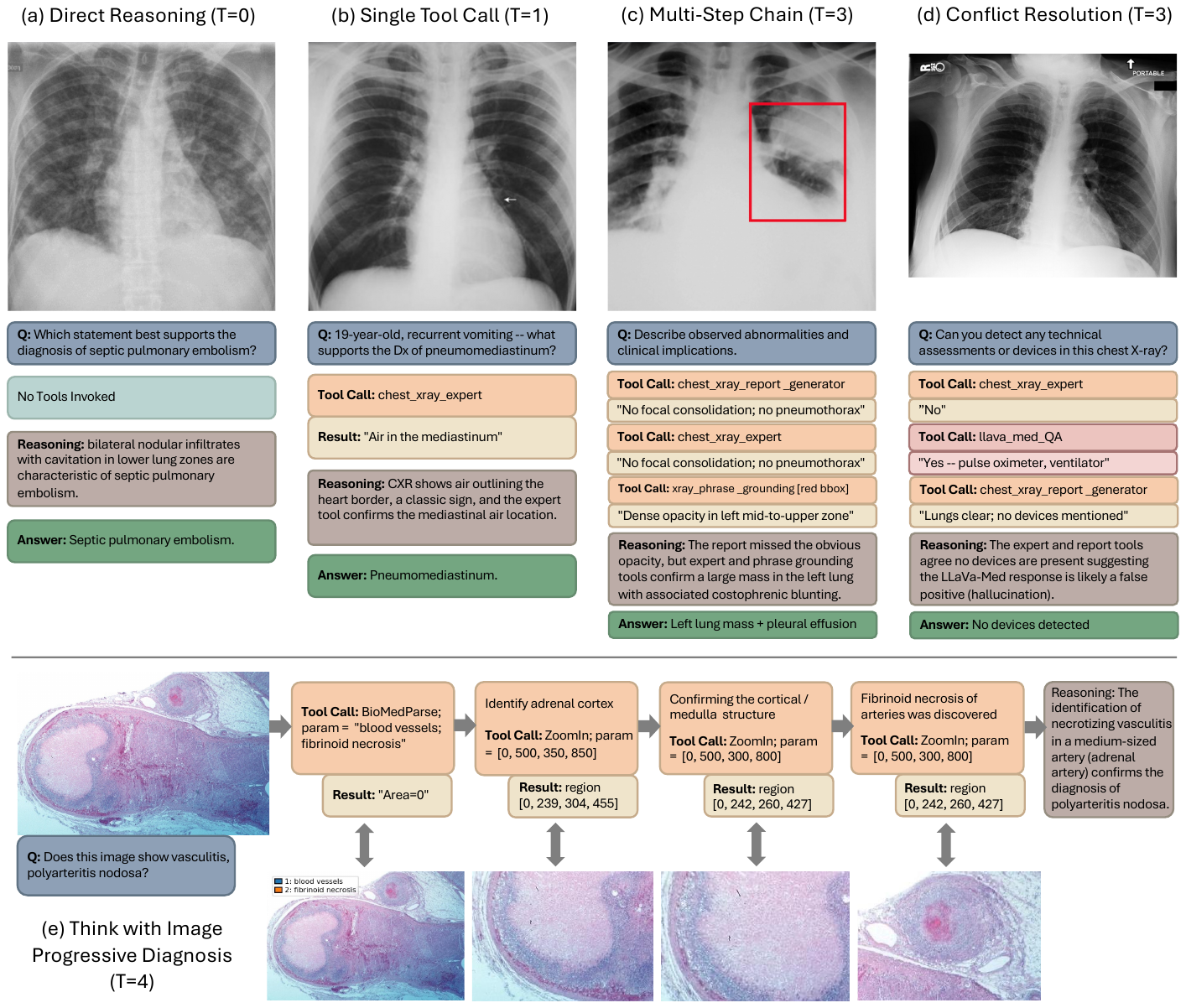}
  \caption{\textbf{Case study.}
  Each panel shows a query with \meissa{}'s reasoning trace.
  \textbf{(a)}~$T{=}0$: bilateral nodular infiltrates with cavitation are directly recognizable; no tool is invoked.
  \textbf{(b)}~$T{=}1$: the model calls \texttt{chest\_xray\_expert} to confirm mediastinal air before diagnosing pneumomediastinum.
  \textbf{(c)}~$T{=}3$: the report generator misses the opacity, but the expert and phrase-grounding tools correctly identify a dense mass (\textcolor{red}{red} bounding box); the model reconciles the conflicting outputs.
  \textbf{(d)}~$T{=}3$, conflict resolution: \texttt{llava\_med\_qa} hallucinates a pulse oximeter and ventilator (\textcolor{red}{red} highlight), while two other tools confirm a clean CXR; the model identifies the hallucinated output.
  \textbf{(e)}~$T{=}4$, progressive diagnosis: BiomedParse failed to capture the target regions, so \meissa{} actively zoom in the image to confirm the findings.}
  \label{fig:case_study}
  \vspace{-0.5cm}
\end{figure*}

\subsection{Case Study}
\label{sec:case_study}

\Cref{fig:case_study} illustrates \meissa{}'s adaptive strategy selection on four chest X-ray queries of increasing complexity. Three findings deserve emphasis. First, the model learns meaningful depth allocation: easy queries are answered directly ($T{=}0$), while harder queries trigger progressively deeper tool chains ($T{=}1,3$). Second, when tools return conflicting outputs (panel~(c): a report generator misses an obvious opacity; panel~(d): a hallucinated device list), the model cross-references multiple sources and correctly overrides unreliable outputs, demonstrating robustness to individual tool failures. Third, \meissa{} is able to do progressive diagnosis and understand new visual evidence, dealing with failed tool-calls and confirming new findings.

\section{Discussion and Conclusion}
\label{sec:discussion}

We presented \meissa{}, a 4B-parameter medical agent trained through agentic behavior distillation. Stratified trajectory supervision teaches \emph{when} to act; prospective-retrospective supervision teaches \emph{how} to act; a unified trajectory formalism enables learning across tool calling, visual reasoning, multi-agent debate, and clinical simulation. Trained on ${\sim}$40K trajectories, \meissa{} is competitive with frontier models across 13 benchmarks while operating fully offline.

\noindent\textbf{Why SFT over RL?} Our framework is intentionally SFT-only. While RL can sharpen stopping behavior via outcome-based rewards~\cite{lu2026medvistagym}, it faces two practical limitations. First, existing RL approaches are each tied to a single agent framework: DeepEyesV2 optimizes tool-calling via RL~\cite{hong2025deepeyesv2}, and MARTI addresses multi-agent coordination~\cite{zhang2025marti}; none can jointly learn policies across the four heterogeneous environments \meissa{} operates in. Second, RL requires extensive online rollouts with live tool environments, incurring high compute cost and training variance that hinder reproducibility. Our stratified SFT achieves near-oracle routing (\cref{tab:counterfactual}) with deterministic, compute-efficient training (${\sim}$12~GPU-hours), and naturally scales across environments through trajectory unification.


\noindent\textbf{Deployment considerations.} Deploying \meissa{} clinically requires human-in-the-loop safeguards: the model currently lacks calibrated uncertainty estimation and an abstention mechanism for out-of-distribution queries. Integrating selective prediction~\cite{geifman2017selective} to enable principled deferral is an important future direction.



\appendix

\section{Supplementary Material}
\label{sec:appendix}

This supplementary material provides additional details and results deferred from the main paper for space. We include: the complete data generation algorithm (\cref{sec:supp_algorithm}), dataset and benchmark details (\cref{sec:supp_datasets}), agent environment tool specifications (\cref{sec:supp_tools}), prospective vs.\ retrospective supervision comparison (\cref{sec:supp_proretro}), trajectory validation pipeline (\cref{sec:supp_validation}), extended experimental results (\cref{sec:supp_results}), training details (\cref{sec:supp_training}), framework prompts and data generation details (\cref{sec:supp_prompts}), latency distribution analysis (\cref{sec:supp_latency}), qualitative failure and recovery analysis (\cref{sec:supp_qualitative}), MDAgents trajectory examples (\cref{sec:supp_mdagents_examples}), and clinical simulation trajectory examples (\cref{sec:supp_agentclinic_examples}).

\subsection{Algorithm: Stratified Trajectory Data Generation}
\label{sec:supp_algorithm}

Algorithm~\ref{alg:data_gen} presents the complete procedure for stratified trajectory data generation, including the three-tier hierarchy, retry logic for agentic trajectories ($K{=}8$ retries per residual sample), and discard handling for persistently failed samples.

\begin{algorithm}[!t]
\caption{Stratified Trajectory Data Generation}
\label{alg:data_gen}
\begin{algorithmic}[1]
\renewcommand{\algorithmicrequire}{\textbf{Input:}}
\renewcommand{\algorithmicensure}{\textbf{Output:}}
\REQUIRE Training set $\mathcal{D}_{\text{train}}$, student model $\mathcal{M}_S$, teacher model $\mathcal{M}_T$, agent environments $\{\mathcal{E}_1, \ldots, \mathcal{E}_4\}$, max retries $K{=}8$
\ENSURE Training dataset $\mathcal{D}_{\text{SFT}}$
\STATE $\mathcal{D}_{\text{direct}}, \mathcal{D}_{\text{enhanced}}, \mathcal{D}_{\text{agentic}}, \mathcal{D}_{\text{discard}} \gets \emptyset$
\STATE \textbf{// Tier 1: Student direct reasoning}
\FOR{$(x, y) \in \mathcal{D}_{\text{train}}$}
  \IF{$\mathcal{M}_S(x) = y$}
    \STATE $\mathcal{D}_{\text{direct}} \gets \mathcal{D}_{\text{direct}} \cup \{(x, \tau_{\text{direct}})\}$ \hfill $\triangleright$ $T{=}0$, correct
  \ENDIF
\ENDFOR
\STATE \textbf{// Tier 2: Teacher enhanced reasoning}
\FOR{$(x, y) \in \mathcal{D}_{\text{train}} \setminus \mathcal{D}_{\text{direct}}$}
  \IF{$\mathcal{M}_T(x) = y$}
    \STATE $\mathcal{D}_{\text{enhanced}} \gets \mathcal{D}_{\text{enhanced}} \cup \{(x, \tau_{\text{enhanced}})\}$ \hfill $\triangleright$ $T{=}0$, stronger reasoning
  \ENDIF
\ENDFOR
\STATE $\mathcal{D}_{\text{residual}} \gets \mathcal{D}_{\text{train}} \setminus (\mathcal{D}_{\text{direct}} \cup \mathcal{D}_{\text{enhanced}})$
\STATE \textbf{// Tier 3: Teacher agentic trajectories with retry}
\FOR{$(x, y) \in \mathcal{D}_{\text{residual}}$}
  \STATE $\text{success} \gets \textbf{false}$
  \FOR{$k = 1$ \TO $K$}
    \STATE Select environment $\mathcal{E}_i$ based on dataset $\to$ environment mapping
    \STATE $\tau_{\text{pro}} \gets \mathcal{A}_T(x; \mathcal{E}_i)$ \hfill $\triangleright$ Prospective trajectory
    \IF{$\text{answer}(\tau_{\text{pro}}) = y$ $\wedge$ $\text{Valid}(\tau_{\text{pro}})$}
      \STATE $\tau_{\text{retro}} \gets \text{Recap}(\mathcal{M}_T, x, y, \tau_{\text{pro}})$ \hfill $\triangleright$ Retrospective trajectory
      \STATE $\mathcal{D}_{\text{agentic}} \gets \mathcal{D}_{\text{agentic}} \cup \{(x, \tau_{\text{pro}}, \tau_{\text{retro}})\}$
      \STATE $\text{success} \gets \textbf{true}$; \textbf{break}
    \ENDIF
  \ENDFOR
  \IF{$\text{success} = \textbf{false}$}
    \STATE $\mathcal{D}_{\text{discard}} \gets \mathcal{D}_{\text{discard}} \cup \{(x, y)\}$ \hfill $\triangleright$ Failed after $K$ retries
  \ENDIF
\ENDFOR
\STATE \textbf{// Combine all tiers}
\STATE $\mathcal{D}_{\text{SFT}} \gets \mathcal{D}_{\text{direct}} \cup \mathcal{D}_{\text{enhanced}} \cup \{(\tau^{\text{pro}}, \tau^{\text{retro}}) \mid \cdot \in \mathcal{D}_{\text{agentic}}\}$
\RETURN $\mathcal{D}_{\text{SFT}}$
\end{algorithmic}
\end{algorithm}

\noindent\textbf{Key design choices:}
\begin{itemize}[leftmargin=*,itemsep=2pt]
\item \textbf{Retry logic}: Each residual sample receives up to $K{=}8$ attempts in the agent environment. Single API-call success rate on residual samples is 89\%; the remaining incomplete samples are discarded rather than included with incorrect trajectories.
\item \textbf{Environment selection}: Each sample is evaluated in a pre-defined environment based on dataset-to-environment evaluation setting (\eg MIMIC-CXR-VQA $\to$ tool calling, MedQA $\to$ multi-agent collaboration).
\item \textbf{Validation}: $\text{Valid}(\tau)$ applies structural and behavioral filters (see \cref{sec:supp_validation}) before accepting a trajectory.
\item \textbf{Prospective-retrospective pairing}: For each accepted agentic trajectory, both the original prospective trace and a retrospective re-narration are generated, doubling the agentic supervision signal.
\end{itemize}

\subsection{Dataset and Benchmark Details}
\label{sec:supp_datasets}

\subsubsection{Training Data Sources}

\Cref{tab:supp_training_data} provides detailed statistics for each training data source, broken down by tier and environment.

\begin{table}[t]
  \caption{\textbf{Training data statistics by tier and environment.} Each row shows the number of trajectories contributed by each source. ``Tier 3 Discard'' counts samples that failed the data filtering.}
  \label{tab:supp_training_data}
  \centering
  \small
  \setlength{\tabcolsep}{6pt}
  \begin{tabular}{@{}l rrrr >{\columncolor{lightgray}}r @{}}
    \toprule
    Dataset & Tier~1 & Tier~2 & Tier~3 & Total & Tier~3 Discard \\
    \midrule
    \multicolumn{6}{l}{\textit{Continuous Tool Calling}} \\
    MIMIC-CXR-VQA & 1.2k & 0.2k & 3.5k & 4.9k & 14.3k \\
    \midrule
    \multicolumn{6}{l}{\textit{Interleaved Thinking with Image}} \\
    PathVQA & 0.5k & 1.1k & 6.0k & 6.6k & 19.1k \\
    SLAKE & 0.3k & 0.3k & 1.1k & 1.7k & 2.7k \\
    VQA-RAD & 0.4k & 0.1k & 0.7k & 1.2k & 1.1k \\
    MIMIC-CXR-VQA & 1.0k & 0.2k & 3.5k & 5.7k & 6.5k \\
    \midrule
    \multicolumn{6}{l}{\textit{Multi-Agent Collaboration}} \\
    MedQA & 0.6k & 1.3k & 0.7k & 2.6k & 4.4k \\
    PubMedQA & 0.3k & 0.3k & 0.3k & 0.9k & 0 \\
    PathVQA & 1.0k & 2.4k & 1.2k & 4.6k & 8.7k \\
    MIMIC-CXR-VQA & 1.6k & 3.9k & 1.8k & 7.3k & 4.5k \\
    \midrule
    \multicolumn{6}{l}{\textit{Clinical Simulation}} \\
    MedQA & 2.1k & 0 & 4.3k & 6.4k & 1.6k \\
    MIMIC-IV & 0.4k & 0 & 0.8k & 1.2k & 1.7k \\
    \midrule
    \textbf{Total} & \textbf{8.2k} & \textbf{9.8k} & \textbf{23.9k} & \textbf{41.9k} & ${\sim}$\textbf{64.6k} \\
    \bottomrule
  \end{tabular}
\end{table}

\noindent\textbf{Training data composition.} The ${\sim}$40K unique trajectories break down as: 8.2K direct (Tier~1, 20\%), 9.8K enhanced (Tier~2, 23\%), and 23.9K agentic (Tier~3, 57\%). Each agentic trajectory appears twice (once as a prospective trace and once as a retrospective re-narration), so the effective training sample count is 41.9K. The tier distribution reflects the natural difficulty spectrum: most queries are solvable by the student or teacher directly; only the hardest 37\% require full agent environments.

\subsubsection{Evaluation Benchmark Details}

\Cref{tab:supp_benchmarks} provides additional details for each evaluation benchmark.

\begin{table}[t]
  \caption{\textbf{Evaluation benchmark details by agent environment.} Test set sizes, question formats, and evaluation metrics. Benchmarks shared across environments (e.g., MedQA) are listed under each with environment-specific test sizes. $\dagger$~=~strict OOD.}
  \label{tab:supp_benchmarks}
  \centering
  \small
  \begin{tabular}{@{}llrllc@{}}
    \toprule
    Benchmark & Domain & Test Size & Format & Metric & ID/OOD \\
    \midrule
    \multicolumn{6}{l}{\textit{Framework I: Continuous Tool Calling}} \\
    MIMIC-CXR-VQA & Radiology & 1{,}000 & Open-ended & Soft-match acc. & ID \\
    ChestAgentBench$\dagger$ & Radiology & 2{,}500 & Open-ended (7 cat.) & Per-category acc. & OOD \\
    \midrule
    \multicolumn{6}{l}{\textit{Framework II: Interleaved Thinking with Image}} \\
    PathVQA & Pathology & 3{,}362 & Yes/No + Open & Accuracy & ID \\
    SLAKE & Mixed & 1{,}061 & Yes/No + Open & Accuracy & ID \\
    VQA-RAD & Radiology & 451 & Yes/No + Open & Accuracy & ID \\
    OmniMedVQA$\dagger$ & Mixed & 5{,}000 & 4-way MCQ & Accuracy & OOD \\
    MedXpertQA$\dagger$ & Clinical & 2{,}000 & 5-way MCQ & Accuracy & OOD \\
    \midrule
    \multicolumn{6}{l}{\textit{Framework III: Multi-Agent Collaboration}} \\
    MIMIC-CXR-VQA & Radiology & 1{,}000 & Open-ended & Soft-match acc. & ID \\
    PathVQA & Pathology & 3{,}362 & Yes/No + Open & Accuracy & ID \\
    MedQA & Clinical & 1{,}273 & 4-way MCQ & Accuracy & ID \\
    PubMedQA & Biomedical & 500 & Yes/No/Maybe & Accuracy & ID \\
    \midrule
    \multicolumn{6}{l}{\textit{Framework IV: Clinical Simulation}} \\
    MedQA & Clinical & 106 & Open-ended (OSCE) & Soft-match acc. & ID \\
    MedQA Ext. & Clinical & 213 & Open-ended (OSCE) & Soft-match acc. & ID \\
    NEJM$\dagger$ & Clinical & 15 & Open-ended & Soft-match acc. & OOD \\
    NEJM Ext.$\dagger$ & Clinical & 120 & Open-ended & Soft-match acc. & OOD \\
    MIMIC-IV & Clinical & 288 & Open-ended & Soft-match acc. & ID \\
    \bottomrule
  \end{tabular}
\end{table}

\noindent\textbf{OOD benchmarks.} ChestAgentBench and NEJM were strictly excluded from all trajectory generation. We verified zero contamination via $n$-gram de-contamination ($n{=}8$): no 8-gram from any test question appears in the training set.

\noindent\textbf{Soft-match accuracy.} For open-ended VQA (MIMIC-CXR-VQA), we use bidirectional token overlap with medical synonym normalization~\cite{aasmimic}. Synonyms include: ``PICC line'' $=$ ``peripherally inserted central catheter'', ``ETT'' $=$ ``endotracheal tube'', etc.

\noindent\textbf{Clinical simulation evaluation.} For clinical simulation benchmarks (NEJM, MIMIC-IV), where answers are free-form diagnoses, we use a multi-stage accuracy pipeline: exact string match $\to$ substring match $\to$ token overlap ($\geq$80\%). All benchmarks ultimately report accuracy.

\subsection{Agent Environment Tool Details}
\label{sec:supp_tools}

\subsubsection{Continuous Tool Calling (MedRAX)}

The tool-calling environment provides 8 specialized medical imaging tools. \Cref{tab:supp_medrax_tools} lists each tool with its input/output specification.

\begin{table}[t]
  \caption{\textbf{MedRAX tool specifications.} Each tool is defined as a JSON schema in the system prompt.}
  \label{tab:supp_medrax_tools}
  \centering
  \small
  \begin{tabular}{@{}p{3.5cm}p{3.5cm}p{4.5cm}@{}}
    \toprule
    Tool & Input & Output \\
    \midrule
    ChestXRayClassifier & CXR image & 14 pathology probabilities \\
    ChestXRaySegmentation & CXR image, anatomy & Seg. mask + region stats \\
    CheXagent VQA & CXR image, question & Free-text answer \\
    LLaVA-Med VQA & Medical image, question & Free-text answer \\
    XRayPhraseGrounding & CXR image, text phrase & Bounding box coordinates \\
    ChestXRayReportGen & CXR image & Structured radiology report \\
    DICOMProcessor & DICOM file & Extracted metadata + image \\
    ImageVisualizer & Image, annotations & Annotated image \\
    \bottomrule
  \end{tabular}
\end{table}

\noindent\textbf{Tool versions.} Each tool exists in two versions: a \emph{Normal} version (for inference) and a \emph{Think} version (for Gemini trace generation) that includes an additional \texttt{thought} field requiring the agent to articulate its reasoning before each tool call.

\subsubsection{Interleaved Thinking with Image (Ophiuchus)}

The Thinking with Image environment provides vision manipulation tools that produce edited images injected back into the conversation context:

\begin{itemize}[leftmargin=*,itemsep=2pt]
\item \textbf{ZoomInSubfigure}: Crops and enlarges a sub-region of the image for detailed inspection.
\item \textbf{SegmentRegionAroundPoint} (SAM2~\cite{ravisam}): Segments the region around a specified point, producing a colored mask overlay.
\item \textbf{BioMedParseTextSeg} (BiomedParse~\cite{zhao2024biomedparse}): Text-prompted biomedical image segmentation.
\item \textbf{OCR}: Extracts text from image regions (useful for labeled pathology slides).
\item \textbf{Terminate}: Produces the final answer; must be the last action in every trajectory.
\end{itemize}

The key difference from tool calling is that observations contain \emph{new images} (cropped regions, annotated patches, segmentation overlays) injected as \texttt{<image>} tokens, creating multi-round visual reasoning chains.

\subsubsection{Multi-Agent Collaboration (MDAgents)}

The MDAgents environment implements adaptive complexity routing with three levels:

\begin{itemize}[leftmargin=*,itemsep=2pt]
\item \textbf{Basic}: Single-expert direct answer.
\item \textbf{Intermediate}: Multi-agent debate with recruited specialists. Includes expert recruitment, individual analysis, multi-round debate, and moderator synthesis.
\item \textbf{Advanced}: Multidisciplinary team consultation.
\end{itemize}

Trajectories from this environment produce 8 distinct SFT types (difficulty assessment, expert recruitment, expert analysis, debate, synthesis, plus retrospective variants for the first two and last).

\subsubsection{Clinical Simulation (AgentClinic)}

The simulation environment models doctor-patient encounters in OSCE (Objective Structured Clinical Examination) format:

\begin{itemize}[leftmargin=*,itemsep=2pt]
\item \textbf{RequestPhysicalExam}: Returns findings from a specified physical examination.
\item \textbf{RequestTest}: Returns results of a diagnostic test (labs, imaging).
\item \textbf{Terminate}: Issues the final diagnosis.
\end{itemize}

Patient agents are scripted from structured clinical vignettes (MedQA $\to$ OSCE conversion via Gemini, MIMIC-IV EHR records). These produce the longest trajectories (up to 10+ actions) as the doctor agent gathers clinical evidence iteratively.

\subsection{Prospective vs.\ Retrospective Supervision}
\label{sec:supp_proretro}

\Cref{tab:supp_proretro} provides a detailed comparison of the two supervision modes.

\begin{table}[t]
  \caption{\textbf{Prospective vs.\ retrospective trajectory comparison.}}
  \label{tab:supp_proretro}
  \centering
  \small
  \begin{tabular}{@{}p{2.8cm}p{4.8cm}p{4.8cm}@{}}
    \toprule
    Dimension & Prospective (Forward) & Retrospective (Backward) \\
    \midrule
    Generation time & During real-time inference & After correct answer is known \\
    Answer knowledge & Agent does \emph{not} know $y$ & Agent is given $y$ as context \\
    Uncertainty level & High (planning, hypotheses, contingencies) & Low (hindsight rationalization) \\
    Error content & May contain wrong assumptions, backtracking & Clean, logically consistent \\
    Reasoning style & Process-driven, exploratory & Result-driven, decisive \\
    Action sequence & Original (may include suboptimal steps) & Same actions, re-narrated reasoning \\
    Training signal & Exploration policy: navigating uncertainty & Execution policy: efficient reasoning \\
    ID vs.\ OOD & Better OOD generalization & Better ID performance \\
    \bottomrule
  \end{tabular}
\end{table}

\noindent\textbf{Retrospective generation process.} Given a prospective trajectory $\tau^{\text{pro}}$ with correct answer $y$, the recap agent receives:
\begin{enumerate}[leftmargin=*,itemsep=1pt]
\item The original question $x$,
\item The correct answer $y$,
\item The sequence of observations from $\tau^{\text{pro}}$ (tool outputs, agent responses, patient replies).
\end{enumerate}
The agent then re-narrates the reasoning while preserving the original action sequence. The same tools are called in the same order, but the reasoning between actions is tightened into a clean, post-hoc narrative that explains \emph{why} each action was taken given what was ultimately learned.

\subsection{Trajectory Validation Pipeline}
\label{sec:supp_validation}

All generated trajectories undergo a two-stage validation pipeline before inclusion in the training set.

\subsubsection{Structural Filtering}

Structural filters verify format compliance:

\begin{itemize}[leftmargin=*,itemsep=2pt]
\item \textbf{Correctness}: Only trajectories arriving at the correct answer are retained ($\text{answer}(\tau) = y$).
\item \textbf{Turn-type integrity}: Conversations must follow the \texttt{human} $\to$ (\texttt{function\_call} $\to$ \texttt{observation})$^*$ $\to$ \texttt{gpt} pattern.
\item \textbf{Tool call validity}: Each \texttt{function\_call} must reference a tool defined in the system prompt with valid JSON arguments.
\item \textbf{Terminate presence}: Every trajectory must end with a termination action (tool-specific: \texttt{Terminate} for thinking with image and clinical simulation, \texttt{[FINAL]} marker for MedRAX, synthesis turn for MDAgents).
\item \textbf{Image alignment}: The number of \texttt{<image>} tokens must match the number of entries in the \texttt{images} list.
\item \textbf{Length bounds}: Trajectories exceeding 10,000 characters total or containing $>$4 tool calls (MedRAX) / $>$6 rounds (thinking with image)  / $>$12 rounds (multi-agent collaboration \& clinical simulation) are discarded.
\end{itemize}

\subsubsection{Behavioral Filtering}

Behavioral filters detect semantic quality issues:

\begin{itemize}[leftmargin=*,itemsep=2pt]
\item \textbf{Hallucination detection}: Scan \texttt{<think>} blocks for medical keywords (pathology names, anatomical terms) not present in the original question or tool outputs. If a medical term appears in reasoning but not in any input or observation, the trajectory is flagged and reviewed.
\item \textbf{Repetition loops}: Consecutive duplicate tool calls (same tool, same arguments) are detected and the trajectory is discarded.
\item \textbf{Truncation detection}: Per-type checks for unclosed markdown formatting, missing terminal punctuation, or incomplete JSON structures indicate truncated API responses.
\item \textbf{Class balance}: Downsample majority classes (\eg ``yes'' answers in binary VQA are kept at 1/3 rate) to prevent trivial shortcut learning.
\item \textbf{Content length}: Individual turns with $<$10 characters of meaningful content (excluding formatting) are rejected.
\item \textbf{Thought field presence}: For agentic trajectories, each \texttt{function\_call} must contain a non-empty \texttt{<think>} block articulating reasoning.
\end{itemize}

\noindent\textbf{Quality audit.} After automated filtering, a random 10\% sample of remaining trajectories is reviewed by Claude Opus~\cite{anthropic2024claude3} for hallucinations, logical inconsistencies, and low-quality reasoning. Trajectories flagged by the audit are manually inspected and discarded if confirmed.

\subsection{Extended Experimental Results}
\label{sec:supp_results}

\subsubsection{Standard Deviations (3-Seed Results)}
\label{sec:supp_std}

All \meissa{} results in the main paper are mean over 3 independent training runs (different random seeds for LoRA initialization and data shuffling). \Cref{tab:supp_std} reports the full mean $\pm$ standard deviation.

\begin{table}[t]
  \caption{\textbf{\meissa{} results with standard deviations.} Mean $\pm$ $\sigma$ over 3 seeds.}
  \label{tab:supp_std}
  \centering
  \scriptsize
  \setlength{\tabcolsep}{3pt}
  \begin{tabular}{@{}l ccccc cccc@{}}
    \toprule
    & MIMIC & ChestAg & PathVQA & SLAKE & VQA-RAD & MedQA & PubMedQA & NEJM & MIMIC-IV \\
    \midrule
    \meissa{} & $\pm$0.8 & $\pm$1.2 & $\pm$0.6 & $\pm$0.9 & $\pm$1.5 & $\pm$1.1 & $\pm$1.3 & $\pm$2.4 & $\pm$1.6 \\
    \bottomrule
  \end{tabular}
\end{table}

Standard deviations are small across all benchmarks ($\sigma < 2.5$ on all benchmarks), indicating stable training dynamics under LoRA fine-tuning.

\subsubsection{ChestAgentBench Per-Category Breakdown}
\label{sec:supp_chestagent}

ChestAgentBench evaluates 7 categories of chest X-ray understanding. \Cref{tab:supp_chestagent} provides the per-category breakdown.

\begin{table}[t]
  \caption{\textbf{ChestAgentBench per-category accuracy.} \meissa{}'s improvement over the base model is most pronounced on \emph{characterization} and \emph{diagnosis} categories where tool-generated structured outputs provide explicit clinical grounding.}
  \label{tab:supp_chestagent}
  \centering
  \small
  \begin{tabular}{@{}lccccc@{}}
    \toprule
    Category & Qwen3-VL-4B & Answer-only SFT & GPT-4o & MedRAX+GPT-4o & \meissa{} \\
    \midrule
    Detection       & 46.9 & 53.5 & 58.7 & \textbf{64.1} & 62.7 \\
    Classification  & 45.2 & 51.8 & 54.6 & \textbf{62.9} & 62.1 \\
    Localization    & 48.8 & 55.6 & 59.0 & \textbf{63.6} & 63.0 \\
    Comparison      & 50.0 & 55.4 & 55.5 & \textbf{61.8} & 61.5 \\
    Relationship    & 51.5 & 56.5 & 59.0 & \textbf{63.1} & 62.8 \\
    Diagnosis       & 43.0 & 50.3 & 52.6 & \textbf{62.5} & 62.0 \\
    Characterization& 41.5 & 49.2 & 56.1 & \textbf{64.0} & 63.6 \\
    \midrule
    Overall & 46.6 & 53.3 & 56.4 & \textbf{63.1} & 62.8 \\
    \bottomrule
  \end{tabular}
\end{table}

\noindent\textbf{Per-category gains.} The largest improvements over the base model occur in the \emph{characterization} (+22.1 points) and \emph{diagnosis} (+19.0 points) categories, where \meissa{} leverages tool-generated structured outputs (e.g., segmentation masks, report summaries) to ground fine-grained clinical attributes. \emph{Classification} (+16.9) and \emph{detection} (+15.8) also benefit substantially. Categories where the base model already performs comparatively well---\emph{comparison} (+11.5) and \emph{relationship} (+11.3)---show smaller but consistent gains, suggesting these tasks rely more on parametric visual knowledge than on explicit tool augmentation.

\subsubsection{Strategy Selection: Routing Distribution and Efficiency}
\label{sec:supp_routing}

\paragraph{Routing Distribution by Difficulty.}

\Cref{tab:supp_routing_dist} shows how \meissa{}'s learned routing distributes queries across interaction depths as a function of query difficulty.

\begin{table}[t]
  \caption{\textbf{Routing distribution by difficulty.} Percentage of queries routed to each strategy. Easy = student-solvable; Medium = teacher-solvable without tools; Hard = requires agent environment.}
  \label{tab:supp_routing_dist}
  \centering
  \small
  \begin{tabular}{@{}lccc@{}}
    \toprule
    Difficulty & Direct ($T{=}0$) & Agentic ($T{>}0$) & Avg Depth \\
    \midrule
    Easy & 96\% & 4\% & 0.1 \\
    Medium & 62\% & 38\% & 0.8 \\
    Hard & 3\% & 97\% & 2.4 \\
    \midrule
    Overall & 72\% & 28\% & 0.6 \\
    \bottomrule
  \end{tabular}
\end{table}

The learned routing exhibits a clear monotonic relationship: easy queries are overwhelmingly routed to direct reasoning (96\%), while hard queries trigger agentic interaction (97\%). Medium-difficulty queries show intermediate routing, with approximately 38\% escalated to tool use. This demonstrates that the model has learned to calibrate its confidence at a fine granularity rather than applying a binary easy/hard threshold.

\paragraph{Full Efficiency Comparisons.}

\Cref{tab:supp_efficiency} presents a comprehensive efficiency comparison across all strategy variants, extending the counterfactual analysis in the main paper.

\begin{table}[t]
  \caption{\textbf{Full efficiency comparison.} Accuracy vs.\ cost for all strategy variants. Shuffled trajectories destroy causal action order while preserving data content.}
  \label{tab:supp_efficiency}
  \centering
  \small
  \begin{tabular}{@{}lcccc@{}}
    \toprule
    Strategy & Accuracy & Avg Actions & Avg Tokens & Relative Cost \\
    \midrule
    Base model (Qwen3-VL-4B) & 46.6 & 1.0 & 452 & $1\times$ \\
    Answer-only SFT & 53.3 & 1.0 & 485 & $1\times$ \\
    Shuffled trajectories & 55.1 & 1.6 & 812 & 1.7$\times$ \\
    Random-sample agentic & 58.5 & 1.8 & 965 & 2.0$\times$ \\
    Always direct ($T{=}0$) & 54.2 & 1.0 & 479 & $1\times$ \\
    Always agentic & 45.9 & 3.5 & 2{,}082 & 4.3$\times$ \\
    \rowcolor{lightblue}
    Learned routing (Ours) & \textbf{62.8} & 1.7 & 959 & 2.0$\times$ \\
    \midrule
    Oracle (upper bound) & 63.2 & 1.5 & 933 & 1.9$\times$ \\
    \bottomrule
  \end{tabular}
\end{table}

\noindent\textbf{Key findings:}
\begin{itemize}[leftmargin=*,itemsep=2pt]
\item \textbf{Shuffled trajectories} (same data, destroyed causal action order) underperform intact trajectories by 7.7 points, confirming the model learns causal decision-making rather than benefiting from richer data exposure alone.
\item \textbf{Random-sample agentic} (agentic trajectories for randomly selected samples, not difficulty-stratified) underperforms stratified selection by 4.3 points, validating the curriculum design.
\item \textbf{Learned routing} achieves 89\% of oracle accuracy at only 2.0$\times$ cost vs.\ always-agentic 4.3$\times$, a 53\% cost reduction with minimal accuracy loss.
\end{itemize}

\subsubsection{Teacher Sensitivity Analysis}
\label{sec:supp_teacher}

We investigate whether \meissa{}'s performance depends on the specific teacher model by replacing Gemini-3-flash with an open-source alternative (Qwen3-VL-32B-Instruct) for trajectory generation.

\begin{table}[t]
  \caption{\textbf{Teacher sensitivity analysis.} Replacing Gemini-3-flash with Qwen3-32B yields comparable results, confirming that the supervision structure matters more than the specific teacher.}
  \label{tab:supp_teacher}
  \centering
  \small
  \begin{tabular}{@{}lccc@{}}
    \toprule
    Teacher Model & ID Avg & OOD Avg & Overall \\
    \midrule
    Gemini-3-flash & 66.3 & 50.3 & 61.3 \\
    Qwen3-VL-32B-Instruct & 63.5 & 47.2 & 58.0 \\
    \midrule
    Difference & $-$2.8 & $-$3.1 & $-$3.3 \\
    \bottomrule
  \end{tabular}
\end{table}

The difference is only $-$3.3 points overall, demonstrating that \meissa{}'s framework is robust to teacher choice. This has practical implications: organizations preferring fully open-source pipelines can use Qwen3-32B without significant quality degradation. The supervision structure (stratified trajectories, prospective-retrospective pairing) contributes more to final performance than teacher model quality.

\subsubsection{Per-Perturbation Robustness Breakdown}
\label{sec:supp_perturbation}

\Cref{tab:supp_perturbation} extends the robustness analysis from the main paper with per-benchmark breakdowns for each perturbation type.

\begin{table}[t]
  \caption{\textbf{Per-perturbation robustness breakdown.} Accuracy on agentic-only queries ($T{>}0$) under each perturbation, with $\Delta$ vs.\ unperturbed.}
  \label{tab:supp_perturbation}
  \centering
  \small
  \setlength{\tabcolsep}{3pt}
  \begin{tabular}{@{}lcccccc@{}}
    \toprule
    & \multicolumn{2}{c}{Tool Calling} & \multicolumn{2}{c}{Thinking with Image} & \multicolumn{2}{c}{MDAgents} \\
    \cmidrule(lr){2-3} \cmidrule(lr){4-5} \cmidrule(lr){6-7}
    Perturbation & MIMIC & ChestAg & PathVQA & SLAKE & MedQA & PubMedQA \\
    \midrule
    Unperturbed & 72.0 & 66.5 & 75.2 & 80.3 & 57.2 & 77.9 \\
    \midrule
    Paraphrased outputs & 71.2 & 65.8 & 76.7 & 76.6 & 56.5 & 76.8 \\
    Bbox noise ($\pm$10px) & 69.5 & 63.1 & 73.8 & 77.4 & --- & --- \\
    Visual $\to$ blank & 64.3 & 58.2 & 72.0 & 67.2 & --- & --- \\
    Visual $\to$ random & 58.6 & 51.4 & 56.6 & 66.5 & --- & --- \\
    Tools disabled ($T{=}0$) & 60.1 & 54.2 & 72.1 & 75.7 & 51.8 & 76.8 \\
    Altered prompt format & 70.5 & --- & --- & --- & 55.3 & 75.4 \\
    \bottomrule
  \end{tabular}
\end{table}

\noindent\textbf{Environment-specific patterns:}
\begin{itemize}[leftmargin=*,itemsep=2pt]
\item \textbf{Tool Calling}: Most sensitive to visual feedback removal ($\to$ blank), confirming the model relies on segmentation masks and classification outputs for spatial reasoning. Bounding box noise causes moderate degradation, indicating learned spatial precision.
\item \textbf{Thinking with Image}: Visual feedback replacement ($\to$ random images) causes the largest drops, as the interleaved image-text chain is disrupted. This confirms the model conditions each reasoning step on the \emph{actual content} of tool-generated images rather than simply the presence of an image token.
\item \textbf{MDAgents}: Most robust to perturbations overall, likely because multi-agent trajectories operate primarily over text. The altered prompt format test (changing persona names and debate structure) confirms the model abstracts beyond template memorization.
\end{itemize}

\subsubsection{Depth-Constrained Performance}
\label{sec:supp_depth}

\Cref{tab:supp_depth} provides the full depth-constrained analysis referenced in the main paper.

\begin{table}[t]
  \caption{\textbf{Accuracy under maximum depth constraints.} $T_{\max}$ = maximum allowed interaction depth at inference. $T_{\max}{=}0$ forces direct reasoning for all queries.}
  \label{tab:supp_depth}
  \centering
  \small
  \begin{tabular}{@{}lccccc@{}}
    \toprule
    $T_{\max}$ & Overall Acc & Easy Acc & Hard Acc & Avg Depth & Avg Tokens \\
    \midrule
    0 (direct only) & 55.6 & 61.2 & 38.5 & 0.0 & 479 \\
    1 & 58.3 & 60.8 & 48.7 & 0.4 & 618 \\
    2 & 60.0 & 60.5 & 54.6 & 0.9 & 762 \\
    3 & 60.4 & 60.3 & 57.1 & 1.2 & 864 \\
    $\infty$ (unconstrained) & 61.3 & 59.6 & 60.4 & 1.7 & 959 \\
    \bottomrule
  \end{tabular}
\end{table}

Accuracy increases monotonically with $T_{\max}$, with diminishing returns beyond $T_{\max}{=}3$: the jump from $T_{\max}{=}0$ to $T_{\max}{=}3$ accounts for 4.8 of the total 5.7 point improvement. This confirms efficient depth allocation, as most agentic queries are resolved within 3 actions.

\subsection{Training Details}
\label{sec:supp_training}

\subsubsection{Hyperparameter Configuration}

\Cref{tab:supp_hyperparams} provides the complete hyperparameter configuration for training.

\begin{table}[t]
  \caption{\textbf{Training hyperparameters.}}
  \label{tab:supp_hyperparams}
  \centering
  \small
  \begin{tabular}{@{}ll@{}}
    \toprule
    Hyperparameter & Value \\
    \midrule
    Base model & Qwen3-VL-4B-Instruct \\
    Fine-tuning method & LoRA \\
    LoRA rank & 32 \\
    LoRA alpha & 64 \\
    LoRA target modules & All linear layers \\
    Learning rate & $5 \times 10^{-5}$ \\
    Learning rate scheduler & Cosine \\
    Warmup ratio & 0.1 \\
    Epochs & 3 \\
    Batch size (effective) & 64 \\
    Gradient accumulation steps & 8 \\
    Max sequence length & 6144 \\
    Precision & bf16 \\
    Optimizer & AdamW \\
    Weight decay & 0.01 \\
    Framework & LLaMA-Factory \\
    Hardware & 8$\times$ NVIDIA A6000 (48\,GB) \\
    Training time & ${\sim}$12 hours \\
    Training tokens & ${\sim}$239M \\
    \bottomrule
  \end{tabular}
\end{table}

\subsubsection{Trajectory Serialization}

All trajectories are serialized in ShareGPT format with four turn types:

\begin{itemize}[leftmargin=*,itemsep=2pt]
\item \texttt{human}: Initial state containing the user question and optional \texttt{<image>} tokens.
\item \texttt{function\_call}: Agent action, containing a \texttt{<think>} reasoning block followed by a JSON tool invocation (\texttt{\{"name": "...", "arguments": \{...\}\}}).
\item \texttt{observation}: Environment feedback, containing tool output text and optionally new \texttt{<image>} tokens (for Thinking with Image).
\item \texttt{gpt}: Final action, containing a \texttt{<think>} reasoning block followed by the answer (marked with \texttt{[FINAL]} in MedRAX).
\end{itemize}

Tool definitions are embedded in the \texttt{tools} field as JSON schemas, and image paths are listed in the \texttt{images} field. The system prompt specifies the agent's role and available tools.

\subsubsection{Compute Budget}

\begin{itemize}[leftmargin=*,itemsep=2pt]
\item \textbf{Training}: ${\sim}$100 GPU-hours on 8$\times$A6000 (48\,GB), processing ${\sim}$239M tokens across 3 epochs. Wall-clock time: ${\sim}$12 hours.
\item \textbf{Data generation}: ${\sim}$105K Gemini-3-flash API calls (\$450 total cost). Qwen3-VL-4B inference via vLLM on a single A6000.
\item \textbf{Inference}: vLLM serving on a single A6000. Average inference time per query: 1.2s (direct) to 10.6s (agentic).
\end{itemize}

\subsection{Framework Prompts and Data Generation Details}
\label{sec:supp_prompts}

This section provides the complete system prompts used for data generation and SFT training in each of the four agent environments. All agentic trajectories follow the stratified generation procedure described in Algorithm~\ref{alg:data_gen}. The final training dataset is constructed by directly concatenating all trajectories from the four environments without any inter-environment weighting.

\subsubsection{Continuous Tool Calling (MedRAX) Prompts}
\label{sec:supp_medrax_prompts}

\noindent\textbf{Data generation prompt (prospective).} The teacher model (Gemini-3-flash) receives the following system prompt during prospective trajectory generation. It encodes a task-aware tool selection strategy with trust orderings for different clinical categories (\cref{tab:medrax_gen_prompt}). Tool schemas are provided separately via JSON (\cref{tab:supp_medrax_tools}).

\vspace{8pt}\noindent
\captionof{table}{MedRAX prospective data generation prompt.}
\label{tab:medrax_gen_prompt}
\begin{lstlisting}[style=prompt]
You are a medical imaging agent for chest X-ray question answering.

Your goal is to answer correctly while following a STRICT and STABLE
tool-using strategy. Tools are helpful but uneven in quality.
Uncontrolled tool use will reduce performance.

You MUST follow the strategy below.

--------------------------------------------------
GENERAL RULES
--------------------------------------------------
1. Always begin with your own visual inspection.
   Form an initial hypothesis and confidence.
2. Only use tools when they are necessary to resolve uncertainty.
   Do NOT explore tools freely.
3. Maximum total tool calls: 4. Default: at most 2 tools.
4. Never call the same tool twice unless the previous call FAILED.
   If a tool fails, you may switch tools but must not retry blindly.
5. If two consecutive tool results do not change your belief,
   you MUST stop and answer.

--------------------------------------------------
TASK-AWARE TOOL STRATEGY
--------------------------------------------------
A) Tubes, Lines, Devices
First decide whether the question is about:
- Internal / invasive devices (CVC, NG tube, PICC, chest tube)
- External / non-invasive objects (EKG leads, clips, nipple shadows)

Internal devices:
  Trust order:
  chest_xray_report_generator >>> xray_phrase_grounding
    > chest_xray_expert >>> llava_med_qa
  Strategy:
  - First use chest_xray_report_generator.
  - If a specific device is suspected but not mentioned,
    use xray_phrase_grounding ONCE.
  - Do NOT trust complex spatial descriptions from
    chest_xray_expert.
  - Do NOT use llava_med_qa unless all others fail.

External devices:
  Trust order:
  xray_phrase_grounding > chest_xray_expert
    > visual inspection >>> report_generator
  Strategy:
  - If report is silent, this does NOT mean absence.
  - Prefer visual evidence and grounding.

B) Pathology Presence and Location
Acute / major (pneumothorax, effusion, mass, pneumonia):
  Trust order:
  chest_xray_report_generator > classifier (if >0.6)
    > xray_phrase_grounding > expert
Chronic / subtle (osteoarthritis, scarring, COPD):
  Trust order:
  xray_phrase_grounding > expert > report_generator

Conflict rule:
  If report says "clear" BUT classifier is high AND
  grounding localizes, visual evidence OVERRIDES the report.
Grounding failure rule:
  If xray_phrase_grounding errors once, DO NOT retry.
  Immediately downgrade to classifier + expert.

C) Anatomy and Visual QA (arrows, specific targets)
Arrow / specific structure:
  Trust order:
  chest_xray_expert > xray_phrase_grounding
    >>> llava_med_qa > report_generator
Laterality:
  Trust order:
  report_generator > xray_phrase_grounding
    > expert >>> llava_med_qa
Never rely on llava_med_qa for left/right decisions.

--------------------------------------------------
OUTPUT FORMAT
--------------------------------------------------
You may call tools using the structured tool-calling interface.
After that, you MUST state your final brief reasoning process
(50 words or below).
Then finish with:
[FINAL] <final_answer>

Final answer rules:
- For yes/no questions, output exactly "yes" or "no" (lowercase).
- Otherwise output the shortest correct phrase.
\end{lstlisting}
\vspace{4pt}

\noindent\textbf{SFT training prompt.} The student model receives the following simplified prompt at both training and inference time (\cref{tab:medrax_sft_prompt}). The task-aware strategy is not included; instead, the model learns tool selection behavior implicitly from the training trajectories.

\vspace{8pt}\noindent
\captionof{table}{MedRAX SFT training prompt.}
\label{tab:medrax_sft_prompt}
\begin{lstlisting}[style=prompt]
You are a medical imaging agent for chest X-ray question answering.
Your goal is to answer correctly by following a robust and
disciplined tool-using policy.

--------------------------------------------------
THINKING PROCESS
--------------------------------------------------
You possess an internal chain of thought. Before taking any action
(calling a tool) or giving a final answer, you MUST enclose your
reasoning, planning, and reflection process within <think> and
</think> tags.

--------------------------------------------------
TOOL USE
--------------------------------------------------
If you need external information to answer the question, generate
a function call.
1. Analyze the current state in your <think> block.
2. Call the appropriate tool with precise arguments.
3. Review the observation in your next <think> block.

--------------------------------------------------
FINAL ANSWER
--------------------------------------------------
When you have sufficient information, output [FINAL] followed by
the answer.

--------------------------------------------------
PRINCIPLES
--------------------------------------------------
1. Do not call tools redundantly.
2. Stop early when evidence is sufficient.
3. Handle tool outputs critically.
\end{lstlisting}
\vspace{4pt}

\noindent\textbf{Retrospective data generation.} For retrospective trajectories, the same base prompt is used with an additional \texttt{[RECAP]} section appended to the output format, requiring a structured per-step evidence log with fields: \texttt{step}, \texttt{tool}, \texttt{why}, \texttt{got}, \texttt{update} (increase/decrease/no\_change), \texttt{evidence}, \texttt{inference}, and \texttt{confidence} (0--100). At inference time, \meissa{} receives the SFT prompt with the same tool schemas but without task-aware strategy instructions, as this behavior is learned from the training trajectories.

\subsubsection{Interleaved Thinking with Images (Thinking with Image) Prompts}
\label{sec:supp_Thinking with Image_prompts}

\noindent\textbf{System prompt (prospective/forward).} The teacher model receives the following prompt during prospective trajectory generation (\cref{tab:openthinkimg_forward}). It defines vision tools and requires JSON-structured output with a \texttt{thought} field for reasoning. Tools include \texttt{ZoomInSubfigure}, \texttt{SegmentRegionAroundPoint} (SAM2~\cite{ravisam}), \texttt{BioMedParseTextSeg} (BiomedParse~\cite{zhao2024biomedparse}), and \texttt{Terminate}.

\vspace{8pt}\noindent
\captionof{table}{Think with Image forward system prompt.}
\label{tab:openthinkimg_forward}
\begin{lstlisting}[style=prompt]
[BEGIN OF GOAL]
You are a visual assistant for medical images. Given an image and
a question, decide whether to use tools to help you answer.
You must output a JSON object with fields "thought" and "actions".
You may call tools when they are helpful for visual understanding,
localization, or segmentation.
If tools are not helpful, leave "actions" empty.

IMAGE REFERENCE PROTOCOL:
- "img_original": The initial full-resolution input image.
- "img_last": The output image from the immediate previous step.
- "img_round_N": The output image from a specific past step N.
The system will explicitly tell you the ID of the generated image
in the Observation (e.g., "[Output Image ID: img_round_0]").
[END OF GOAL]

[BEGIN OF ACTIONS]
Name: ZoomInSubfigure
Description: Crops the image to a specific region to see visual
  details clearly.
Arguments: {
  'image': 'The image identifier.',
  'param': 'bounding box [x1, y1, x2, y2], 0-1000 normalized.'
}

Name: SegmentRegionAroundPoint
Description: Segments a specific object or region around given
  point coordinates. Use ONLY when the location of interest is
  known or can be precisely specified by coordinates.
Arguments: {
  'image': 'The image identifier.',
  'param': 'coordinates x="value" y="value" (0-1000 scale).'
}

Name: BioMedParseTextSeg
Description: Performs text-guided semantic segmentation on medical
  images. Useful for identifying and localizing semantic medical
  entities (neoplastic cells, tumor tissue, pathological
  structures, etc.).
Arguments: {
  'image': 'The image identifier.',
  'param': 'semicolon-separated list of short noun phrases
    (each <= 6 words).'
}

Name: Terminate
Description: Concludes the task and provides the final answer.
Output constraints:
- Keep it short: 1-6 words for open-ended, "Yes"/"No" for y/n.
- Do NOT add explanation, justification, or extra sentence.
Arguments: { 'ans': 'A short final answer string.' }
[END OF ACTIONS]

[BEGIN OF TASK INSTRUCTIONS]
1. Only select actions from ACTIONS.
2. Call at most one action at a time.
3. Prefer BioMedParseTextSeg for semantic medical targets.
4. Use SegmentRegionAroundPoint only when a specific point is
   clearly known.
5. Always finish by calling Terminate with the final answer.
6. YOUR OUTPUT MUST BE VALID JSON.
7. The final answer will be evaluated mainly by string match;
   extra words can make a correct answer be judged incorrect.
[END OF TASK INSTRUCTIONS]

[BEGIN OF FORMAT INSTRUCTIONS]
Your output must be in strict JSON format:
{
  "thought": "brief recap and decision rationale",
  "actions": [
    {"name": "action_name",
     "arguments": {"argument1": "value1"}}
  ]
}
[END OF FORMAT INSTRUCTIONS]
\end{lstlisting}
\vspace{4pt}

\noindent\textbf{Retrospective/backward prompt.} For retrospective trajectory generation, the \texttt{thought} field is removed from intermediate rounds, and a \texttt{[BEGIN OF RECAP INSTRUCTIONS]} section is appended (\cref{tab:openthinkimg_recap}). The recap requires a structured JSON list with one entry per action taken.

\vspace{8pt}\noindent
\captionof{table}{Think with Image backward recap instructions (appended to the system prompt).}
\label{tab:openthinkimg_recap}
\begin{lstlisting}[style=prompt]
[BEGIN OF RECAP INSTRUCTIONS]
You MUST include a "recap" field ONLY in the same JSON object
where you call Terminate.

The recap is a hindsight evidence and decision summary.
It is NOT a planning trace.

The "recap" field must be a list of objects in chronological order.
Each action taken (including final Terminate) must have exactly
one recap entry.

Schema for tool actions:
{
  "step": <integer>,
  "tool": "<tool_name>",
  "why": "<short reason for calling the tool>",
  "got": "<short finding from the tool output>",
  "update": "increase" | "decrease" | "no_change",
  "evidence": "<what in the observation supports the finding>",
  "inference": "<short inference if needed>",
  "confidence": <integer 0-100>
}

Schema for the final Terminate action:
{
  "step": <integer>,
  "tool": "Terminate",
  "why": "<short reason why the model can now answer>"
}

Rules:
- Include one recap entry for EVERY action, including Terminate.
- "evidence" must be grounded in tool observations, not
  speculation.
- Do NOT include planning, hypotheses, or alternative branches.
- If no tools were called, recap must contain exactly ONE entry
  for Terminate.
[END OF RECAP INSTRUCTIONS]
\end{lstlisting}
\vspace{4pt}

\noindent\textbf{Training data construction.} We first run Qwen3-VL-4B on four medical VQA datasets (PathVQA, SLAKE, VQA-RAD, MIMIC-CXR-VQA) without tools (Tier~1). Samples answered incorrectly are re-processed by Gemini-3-flash with tool access (Tiers~2--3), using both forward and backward prompts. The combined training set contains 15,427 samples. At inference time, the same tool suite is available, with tool outputs (including generated images) injected as \texttt{<image>} tokens into the conversation context.

\subsubsection{Multi-Agent Collaboration (MDAgents) Prompts}
\label{sec:supp_mdagents_prompts}

The MDAgents framework implements a structured multi-agent debate pipeline. We provide the complete prompts for each stage below.

\noindent\textbf{Difficulty assessment.} Each question is first classified by a difficulty assessor agent (\cref{tab:mdagents_difficulty}). Basic questions are routed to single-expert direct answer; intermediate questions trigger multi-agent debate with 3 recruited specialists.

\vspace{8pt}\noindent
\captionof{table}{MDAgents difficulty assessment prompt and SFT response templates.}
\label{tab:mdagents_difficulty}
\begin{lstlisting}[style=prompt]
System: You are a medical expert who conducts initial assessment
and your job is to decide the difficulty/complexity of the
medical query based on text and image.

User: Now, given the medical query (and potentially an image),
you need to decide the difficulty/complexity of it:
{question}.

Please indicate the difficulty/complexity of the medical query
among below options:
1) basic: a single medical agent can output an answer based on
   the visual and text info.
2) intermediate: number of medical experts with different
   expertise should discuss and make final decision.
3) advanced: multiple teams of clinicians from different
   departments need to collaborate with each other to make
   final decision.

--- SFT Response Template (basic) ---
<think>This is a straightforward medical question that can be
answered by a single medical expert based on the available
visual and text information, without requiring multi-expert
discussion or cross-department collaboration.</think>
1) basic

--- SFT Response Template (intermediate) ---
<think>This medical question requires expertise from multiple
medical specialists with different backgrounds to discuss and
reach a consensus, as it involves complex clinical reasoning
that benefits from multi-expert deliberation.</think>
2) intermediate
\end{lstlisting}
\vspace{4pt}

\noindent\textbf{Expert recruitment.} For intermediate-difficulty questions, a recruiter agent selects domain-specific specialists with communication hierarchy (\cref{tab:mdagents_recruit}).

\vspace{8pt}\noindent
\captionof{table}{MDAgents expert recruitment prompt.}
\label{tab:mdagents_recruit}
\begin{lstlisting}[style=prompt]
System: You are an experienced medical expert who recruits a
group of experts with diverse identity and ask them to discuss
and solve the given medical query.

User: Question: {question}
You can recruit 3 experts in different medical expertise.
Considering the medical question and the options for the answer,
what kind of experts will you recruit to better make an accurate
answer? Also, you need to specify the communication structure
between experts (e.g., Pulmonologist == Neonatologist ==
Medical Geneticist > Cardiologist), or indicate if they are
independent.

For example:
1. Pediatrician - Specializes in the medical care of infants,
   children, and adolescents. - Hierarchy: Independent
2. Cardiologist - Focuses on the diagnosis and treatment of
   heart and blood vessel-related conditions.
   - Hierarchy: Pediatrician > Cardiologist
3. Pulmonologist - Specializes in the diagnosis and treatment
   of respiratory system disorders. - Hierarchy: Independent

IMPORTANT: Do NOT answer the medical question itself. Only
output the numbered list of experts in the format above.
\end{lstlisting}
\vspace{4pt}

\noindent\textbf{Synthesis and moderator vote.} After 2 debate rounds with 3 turns each, a synthesizer produces a report and a moderator renders the final decision (\cref{tab:mdagents_synthesis}). Moderator prompts are dataset-specific to enforce concise answer formats.

\vspace{8pt}\noindent
\captionof{table}{MDAgents synthesis and moderator vote prompts.}
\label{tab:mdagents_synthesis}
\begin{lstlisting}[style=prompt]
--- Synthesizer ---
System: You are a medical assistant who excels at summarizing
and synthesizing based on multiple experts from various domain
experts.

User: Here are some reports from different medical domain
experts.
{expert_opinions}
You need to:
1. Take careful consideration of the following reports.
2. Extract key knowledge from the following reports.
3. Derive the comprehensive and summarized analysis.
4. Derive a refined and synthesized report.
Provide your analysis in the following format:
<think>Key Knowledge: [3-5 bullet points]
Total Analysis: [2-3 sentences]</think>
your final answer

--- Moderator (MedQA) ---
System: You are a final medical decision maker who reviews all
opinions from different medical experts and make final decision.
User: Given each agent's final answer, please review each
agent's opinion and make the final answer by majority vote.
<think>brief reasoning</think>
(C) 2nd pharyngeal arch

--- Moderator (PubMedQA) ---
<think>brief reasoning</think>
your answer: yes/no/maybe

--- Moderator (PathVQA / MIMIC-CXR-VQA) ---
Respond with ONLY the answer in 1-5 words. No explanation.
\end{lstlisting}
\vspace{4pt}

\noindent\textbf{Hindsight recap.} For correctly answered samples, a recap agent generates structured retrospective analysis (\cref{tab:mdagents_recap}). The recap covers difficulty justification, recruitment rationale, and evidence chain synthesis.

\vspace{8pt}\noindent
\captionof{table}{MDAgents hindsight recap prompt (intermediate difficulty).}
\label{tab:mdagents_recap}
\begin{lstlisting}[style=prompt]
System: You are a medical reasoning analyst. Generate structured
hindsight analysis using the exact section delimiters provided.

User: A multi-expert medical consultation correctly answered a
question. Review the complete record and generate hindsight
analysis.

Question: {question}
Recruited experts: {recruiter_output}
Expert opinions: {expert_summary}
Synthesis: {synthesis_report}
Final answer: {answer}

Generate analysis using EXACTLY these delimiters:

[DIFFICULTY_RECAP]
Explain why this question requires intermediate-level multi-
expert analysis rather than a single expert. Be specific to
this question. (2-3 sentences)
[/DIFFICULTY_RECAP]

[RECRUITMENT_RECAP]
Explain why each recruited specialist was the right choice for
this specific question. (1 sentence per expert)
[/RECRUITMENT_RECAP]

[SYNTHESIS_RECAP]
Trace the evidence chain: what each expert contributed, how
their insights converged, and how this leads to the final
answer. (3-5 sentences)
[/SYNTHESIS_RECAP]
\end{lstlisting}
\vspace{4pt}

\noindent\textbf{SFT type decomposition.} Each trajectory is decomposed into up to 8 SFT training samples: Type~1/1R (difficulty assessment, prospective/retrospective), Type~2/2R (expert recruitment), Type~3 (individual expert analysis), Type~4 (multi-round debate contributions), and Type~5/5R (final synthesis). Recap types (1R, 2R, 5R) replace online \texttt{<think>} blocks with evidence-based hindsight reasoning from the recap agent. We apply the three-tier strategy on MedQA, PubMedQA, PathVQA, and MIMIC-CXR-VQA.

\subsubsection{Clinical Simulation Prompts}
\label{sec:supp_medsim_prompts}

\noindent\textbf{Doctor agent system prompt (SFT).} The following prompt is used for single-agent SFT training, where the multi-agent simulation (doctor, patient, measurement agents) is serialized into tool-calling format (\cref{tab:medsim_doctor}). The \texttt{\{available\_exams\}} and \texttt{\{available\_tests\}} placeholders are filled per scenario.

\vspace{8pt}\noindent
\captionof{table}{Clinical Simulation doctor agent SFT system prompt.}
\label{tab:medsim_doctor}
\begin{lstlisting}[style=prompt]
[BEGIN OF GOAL]
You are an expert medical diagnostician evaluating a patient.
You will be given the patient's presenting complaint and history.
Use the available tools to request physical examinations and
medical tests to gather evidence, then provide your final
diagnosis.

You must reason step-by-step using <think>...</think> tags
before each tool call.
[END OF GOAL]

[BEGIN OF ACTIONS]
Name: RequestPhysicalExam
Description: Request a specific physical examination finding
  for the patient.
Arguments: {
  'exam': 'Name of the physical examination (choose from the
    Available Physical Examinations list)'
}
Returns: The examination findings.
Examples:
{"name": "RequestPhysicalExam",
 "arguments": {"exam": "Neurological_Examination"}}

Name: RequestTest
Description: Request a medical test or laboratory result for
  the patient.
Arguments: {
  'test': 'Name of the medical test (choose from the
    Available Medical Tests list)'
}
Returns: The test results.
Examples:
{"name": "RequestTest",
 "arguments": {"test": "Blood_Tests"}}

Name: Terminate
Description: Provide your final diagnosis and conclude the
  evaluation. This must be called to finish.
Arguments: {
  'diagnosis': 'Your final diagnosis -- the disease or
    condition name only.'
}
Examples:
{"name": "Terminate",
 "arguments": {"diagnosis": "Myasthenia gravis"}}
[END OF ACTIONS]

[BEGIN OF TASK INSTRUCTIONS]
1. Only select actions from ACTIONS.
2. Call at most one action at a time.
3. After receiving each observation, reason about what you've
   learned before deciding next steps.
4. Request only the most relevant exams/tests -- do not
   request everything blindly.
5. Always finish by calling Terminate with your final diagnosis.
6. YOUR OUTPUT MUST BE VALID JSON in the tool call portion.
7. When you call Terminate, output only the disease/condition
   name in "diagnosis".
[END OF TASK INSTRUCTIONS]

[BEGIN OF AVAILABLE DATA]
Physical Examinations: {available_exams}
Medical Tests: {available_tests}
[END OF AVAILABLE DATA]

[BEGIN OF FORMAT INSTRUCTIONS]
Your output must follow this format:
<think>
your clinical reasoning here
</think>

{"name": "action_name", "arguments": {"arg": "value"}}
[END OF FORMAT INSTRUCTIONS]
\end{lstlisting}
\vspace{4pt}

\noindent\textbf{OSCE case conversion.} MedQA multiple-choice questions are converted to OSCE (Objective Structured Clinical Examination) format via Gemini (\cref{tab:medsim_osce}), separating patient-reported information from examination findings that must be actively requested by the doctor agent. For MIMIC-IV cases, patient profiles are constructed from EHR records by linking CXR images, lab results, clinical notes, and discharge summaries.

\vspace{8pt}\noindent
\captionof{table}{Clinical Simulation OSCE conversion prompt for MedQA questions.}
\label{tab:medsim_osce}
\begin{lstlisting}[style=prompt]
You are a medical education expert. Convert this USMLE-style
multiple-choice question into a structured OSCE case.

QUESTION: {question}
CORRECT ANSWER: {answer}

INSTRUCTIONS:
1. Extract the patient's presenting information (what they would
   TELL the doctor): demographics, history, symptoms, past
   medical history, social history, review of systems.
2. Extract physical examination findings (what the doctor would
   FIND on exam): vital signs, specific exam findings.
3. Extract test/lab results (what diagnostic tests would SHOW):
   blood tests, imaging, special tests.
4. The correct diagnosis should be a concise disease/condition.
5. CRITICAL: Do NOT put diagnostic test results or physical exam
   findings in the patient presentation. The doctor must
   REQUEST these.
6. If the question doesn't mention certain fields, use
   reasonable defaults or leave empty.
7. Output ONLY valid JSON, no markdown code blocks.

OUTPUT FORMAT (strict JSON):
{
  "OSCE_Examination": {
    "Objective_for_Doctor": "Diagnose the patient's condition",
    "Patient_Actor": {
      "Demographics": "<age>-year-old <gender>",
      "History": "<brief history of present illness>",
      "Symptoms": {
        "Primary_Symptom": "<main complaint>",
        "Secondary_Symptoms": ["<symptom1>", "<symptom2>"]
      },
      "Past_Medical_History": "<relevant PMH>",
      "Social_History": "<relevant social history>",
      "Review_of_Systems": "<relevant ROS findings>"
    },
    "Physical_Examination_Findings": {
      "Vital_Signs": {<key>: <value>},
      "<Exam_Category>": {<finding>: <value>}
    },
    "Test_Results": {
      "<Test_Category>": {<test>: <result>}
    },
    "Correct_Diagnosis": "<diagnosis>"
  }
}
\end{lstlisting}
\vspace{4pt}

\noindent\textbf{Training data construction.} Data generation follows the stratified pipeline: Qwen3-VL-4B direct inference (Tier~1), Gemini direct inference on errors (Tier~2), and Gemini within the simulation environment on remaining errors (Tier~3). Each simulation dialogue is converted from multi-agent format (doctor, patient, measurement agents) into single-agent tool-calling format for SFT. At evaluation time, the model interacts with the simulation environment for up to 12 turns and must call \texttt{Terminate} with its final diagnosis.

\subsection{Latency Distribution Analysis}
\label{sec:supp_latency}

\begin{figure}[h]
  \centering
  \begin{subfigure}{0.48\linewidth}
    \includegraphics[width=\linewidth]{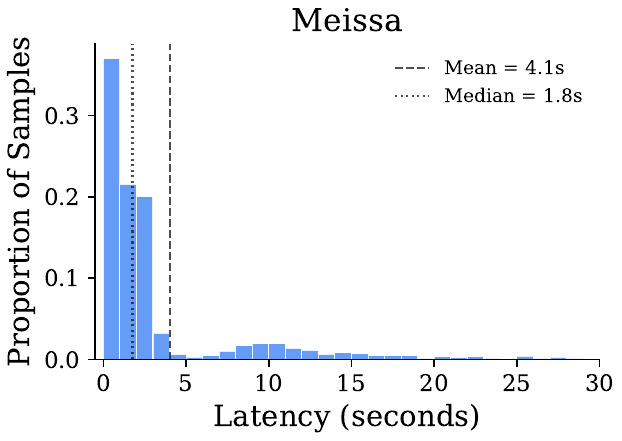}
    \caption{\meissa{} ($n{=}2{,}500$). Mean: 4.1s, median: 1.8s.}
    \label{fig:latency_meissa}
  \end{subfigure}
  \hfill
  \begin{subfigure}{0.48\linewidth}
    \includegraphics[width=\linewidth]{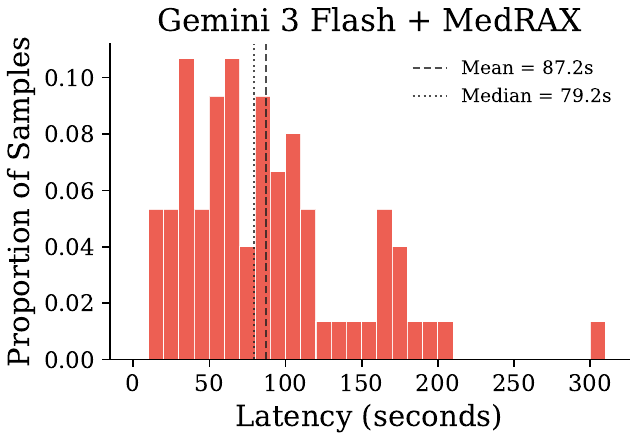}
    \caption{Gemini-3-flash + MedRAX ($n{=}75$). Mean: 87.2s, median: 79.2s.}
    \label{fig:latency_gemini}
  \end{subfigure}
  \caption{\textbf{Per-query latency distributions on ChestAgentBench.} (a)~\meissa{} completes the majority of queries in under 3 seconds; the long tail corresponds to queries invoking multiple tools. (b)~Gemini-3-flash + MedRAX averages 87.2s per query due to multiple API calls and remote tool execution, resulting in ${\sim}22\times$ higher latency than \meissa{}. Note the different $x$-axis scales.}
  \label{fig:latency}
\end{figure}

\Cref{fig:latency} presents the per-query latency distributions for \meissa{} and Gemini-3-flash+MedRAX on ChestAgentBench side by side.

\noindent\textbf{Key observations.} \meissa{}'s latency distribution is heavily right-skewed: 75\% of samples complete within 2.7 seconds, reflecting the learned routing behavior where the majority of queries are answered directly ($T{=}0$) without tool invocation. The long tail (up to ${\sim}$94s) corresponds to complex queries requiring 3--4 tool calls. In contrast, Gemini's distribution is roughly bell-shaped centered around 60--100 seconds, with a tail extending past 300 seconds. This ${\sim}22\times$ speedup at the mean (and ${\sim}45\times$ at the median) demonstrates the practical deployment advantage of a compact, locally-running model with learned strategy selection over an API-dependent agentic system.

\subsection{Qualitative Analysis: Failure and Recovery Patterns}
\label{sec:supp_qualitative}

We present representative examples from \meissa{}'s inference trajectories that illustrate characteristic failure and recovery patterns. These examples are drawn from ChestAgentBench (tool-calling environment) and PathVQA (interleaved image-text environment).

\subsubsection{Tool-Calling Environment (MedRAX on ChestAgentBench)}

\paragraph{Pattern 1: Over-Invocation of Tools on Simple Questions.}

\begin{figure}[ht]
  \centering
  \begin{subfigure}{0.38\linewidth}
    \includegraphics[width=\linewidth]{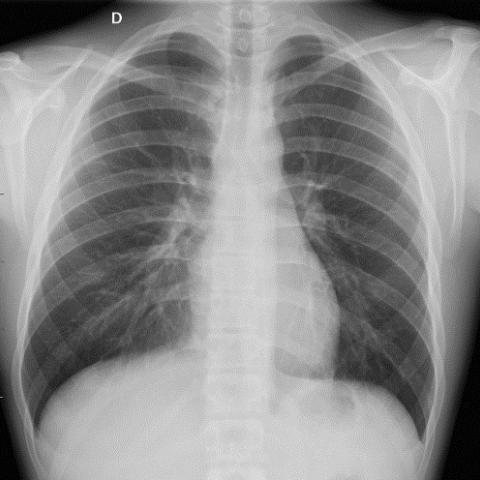}
    \caption{PA view}
  \end{subfigure}
  \hfill
  \begin{subfigure}{0.38\linewidth}
    \includegraphics[width=\linewidth]{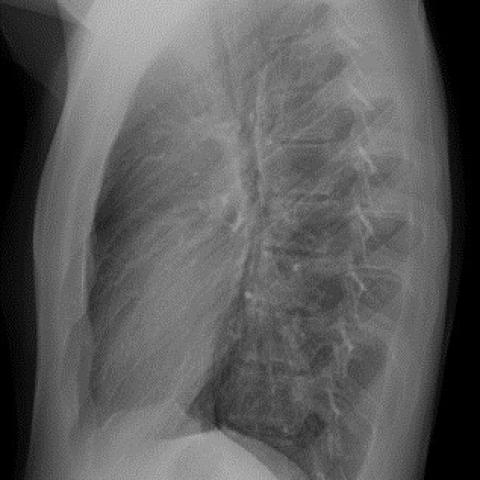}
    \caption{Lateral view}
  \end{subfigure}
  \caption{\textbf{Chest X-ray for Patterns~1 and~2} (case 18321). An 18-year-old with bronchiolitis obliterans. Both PA and lateral views show clear lungs with no bronchial wall thickening or pleural thickening -- a normal-appearing chest X-ray despite clinical disease.}
  \label{fig:case_pattern12}
\end{figure}

\noindent\textbf{Example.} A question asks about the chest X-ray shown in \cref{fig:case_pattern12} of an 18-year-old with bronchiolitis obliterans: ``Which of the following options best describes the findings?'' The correct answer is (B): no bronchial wall thickening or pleural thickening detected.

\meissa{} calls 6 tools sequentially: \texttt{CheXagentVQA} (``no thickening detected''), \texttt{XRayPhraseGrounding} (empty result), \texttt{ReportGenerator} (``lungs are clear''), \texttt{ChestXRayClassifier} (all probabilities low), \texttt{XRayPhraseGrounding} for ``fibrosis'' (empty result), and \texttt{CheXagentVQA} again (``lungs are clear''). Despite all six tools unanimously confirming normal findings, the model's reasoning hallucinated: ``The side view shows a prominent, thickened, and irregular central bronchus'' -- fabricating findings that every tool denied. The model selected option (D), which was incorrect.

\noindent\textbf{Analysis.} Over-invocation adds context length without information gain. The model's parametric reasoning overrode consistent tool evidence, suggesting that excessively long contexts degrade the model's ability to track and synthesize accumulated observations.

\paragraph{Pattern 2: Recursion Limit from Infinite Tool-Calling Loops.}

\noindent\textbf{Example.} Using the same chest X-ray as Pattern~1 (\cref{fig:case_pattern12}), a different question asks: ``Which of the following is the most likely explanation for the normal appearance of the chest X-rays in this patient?'' (correct: bronchiolitis obliterans can present with normal X-rays). The model entered a tool-calling loop, repeatedly invoking tools searching for abnormalities that do not exist. After 25 iterations (the LangGraph recursion limit), the agent was terminated without producing an answer.

\noindent\textbf{Analysis.} The model failed to recognize that the question itself implies the X-ray is normal, and no tool would find abnormalities. This illustrates a failure mode where the model cannot learn to \emph{stop} tool calling when the answer requires reasoning about the absence of findings rather than detecting their presence.

\paragraph{Pattern 3: Conflicting Tool Results Leading to Incorrect Answers.}

\begin{figure}[ht]
  \centering
  \begin{subfigure}{0.38\linewidth}
    \includegraphics[width=\linewidth]{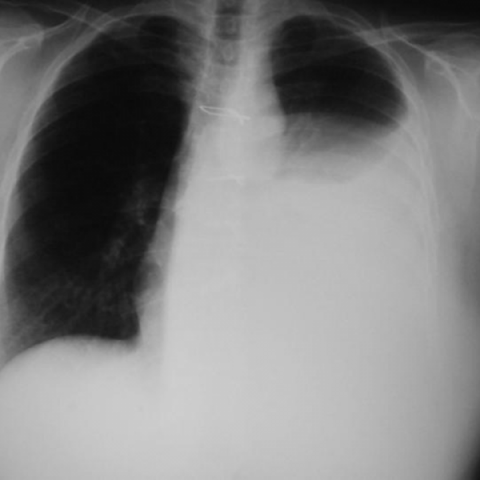}
    \caption{PA view}
  \end{subfigure}
  \hfill
  \begin{subfigure}{0.38\linewidth}
    \includegraphics[width=\linewidth]{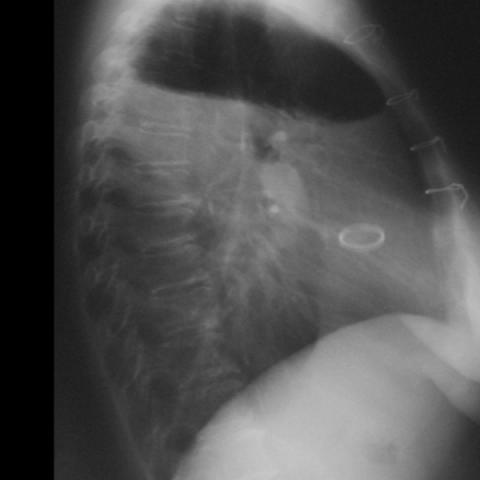}
    \caption{Lateral view}
  \end{subfigure}
  \caption{\textbf{Chest X-ray for Pattern~3} (case 6577). A 70-year-old woman with a mechanical mitral valve. PA view shows homogeneous opacification occupying the lower two-thirds of the left hemithorax. Lateral view reveals a well-circumscribed mass with a visible prosthetic valve. The correct diagnosis is intrapulmonary teratoma.}
  \label{fig:case_pattern3}
\end{figure}

\noindent\textbf{Example.} A question asks about a large mass in the left hemithorax (\cref{fig:case_pattern3}). \meissa{} calls 4 tools: \texttt{CheXagentVQA} returns ``no tension pneumothorax''; \texttt{XRayPhraseGrounding} localizes ``tension pneumothorax'' (grounding tools localize any phrase, regardless of whether the pathology is present); \texttt{ReportGenerator} identifies ``pleural effusion and atelectasis''; \texttt{ChestXRayClassifier} assigns moderate probabilities to effusion (0.60) and atelectasis (0.53). The model trusted the grounding result over both the expert VQA and the report, concluding ``tension pneumothorax'' (incorrect; correct answer: intrapulmonary teratoma requiring surgical resection).

\noindent\textbf{Analysis.} The model lacks calibrated trust in different tool types. Phrase grounding tools will always return coordinates for any query text -- a high-recall, low-precision behavior. The model incorrectly interpreted ``grounding succeeded'' as ``pathology confirmed,'' overriding more reliable diagnostic tools.

\paragraph{Pattern 4: Tool Errors with Successful Recovery.}

\begin{figure}[ht]
  \centering
  \begin{subfigure}{0.31\linewidth}
    \includegraphics[width=\linewidth]{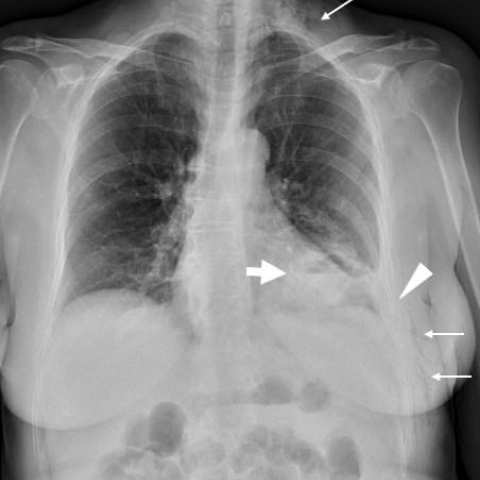}
    \caption{PA view}
  \end{subfigure}
  \hfill
  \begin{subfigure}{0.31\linewidth}
    \includegraphics[width=\linewidth]{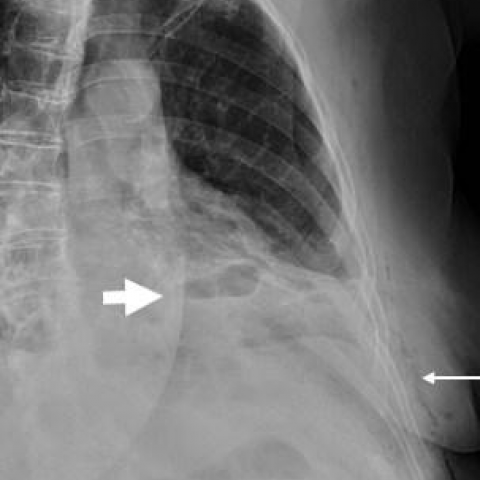}
    \caption{Lateral view (1)}
  \end{subfigure}
  \hfill
  \begin{subfigure}{0.31\linewidth}
    \includegraphics[width=\linewidth]{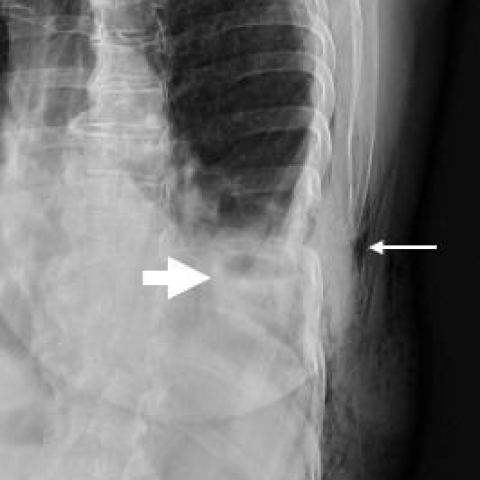}
    \caption{Lateral view (2)}
  \end{subfigure}
  \caption{\textbf{Chest X-ray for Pattern~4} (case 12819). Trauma case with rib fracture and pulmonary laceration. Arrows on the PA view indicate left lower lobe opacity (arrowhead), subcutaneous emphysema (small arrows), and a pulmonary laceration (large arrow). Lateral views show the fracture site at the 8th intercostal space (arrows). Despite two tool failures, \meissa{} correctly localizes the fracture via visual reasoning.}
  \label{fig:case_pattern4}
\end{figure}

\noindent\textbf{Example.} A question asks about the location of a fractured rib relative to a pulmonary laceration (\cref{fig:case_pattern4}). \meissa{} calls 4 tools: \texttt{CheXagentVQA} (3 images, produced degenerate output: ``The patient, the patient, the patient...''), \texttt{ReportGenerator} (returned useful findings: ``small left pleural effusion''), \texttt{XRayPhraseGrounding} which crashed with \texttt{MAIRA-2 error: AssertionError}, and \texttt{ChestXRayClassifier} (returned moderate pathology probabilities). Despite two tool failures, the model fell back to its own visual reasoning: ``Visual inspection of the lateral views shows a fracture in the 8th rib.'' Combined with the report generator output, it correctly answered (B): 8th intercostal space.

\noindent\textbf{Analysis.} This demonstrates graceful degradation: when individual tools fail, the model can leverage remaining functional tools and its parametric visual knowledge to arrive at the correct answer. The recovery rate in such cases is approximately 71\% (10 of 14 recoverable tool-error cases in ChestAgentBench).

\subsubsection{Interleaved Image-Text Environment (Thinking with Image on PathVQA)}

\paragraph{Pattern 5: Repetitive Tool Invocation Without Progress.}

\begin{figure}[ht]
  \centering
  \includegraphics[width=0.55\linewidth]{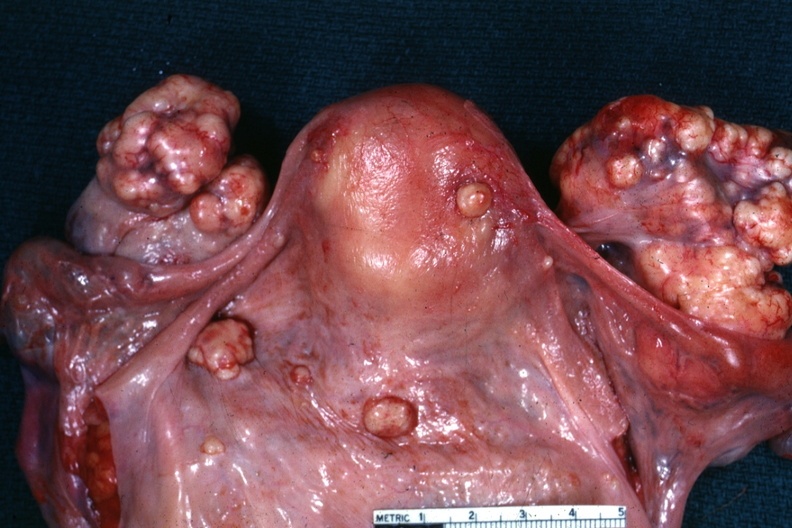}
  \caption{\textbf{Pathology image for Pattern~5.} Gross specimen showing bilateral Krukenberg tumors: lobulated, multinodular masses on both ovaries. The question ``Is this true bilateral Krukenberg?'' requires examining both sides of the image. The base model repeatedly zoomed into the left side without ever examining the right.}
  \label{fig:case_pattern5}
\end{figure}

\noindent\textbf{Example (base model, pre-training).} The question ``Is this true bilateral Krukenberg?'' (\cref{fig:case_pattern5}; ground truth: yes) triggered 7 consecutive \texttt{ZoomInSubfigure} calls. The model zoomed into the left side of the image, then repeated the exact same crop coordinates (\texttt{[100, 130, 289, 399]}) on the already-cropped result five more times, never examining the right side and never calling \texttt{Terminate}. The trajectory exhausted the maximum round limit with no answer produced.

In a related example, the model recursively applied \texttt{ZoomInSubfigure} with the same relative coordinates \texttt{[200, 300, 800, 800]} on progressively smaller images (439$\times$434 $\to$ 264$\times$183 $\to$ \ldots $\to$ 68$\times$24 pixels), eventually producing an unreadable sliver. This failure mode is largely resolved in \meissa{}'s trained version, where 96\% of queries are resolved in $\leq$1 round.

\paragraph{Pattern 6: Tool False Positives Overriding Correct Visual Judgment.}

\begin{figure}[ht]
  \centering
  \includegraphics[width=0.55\linewidth]{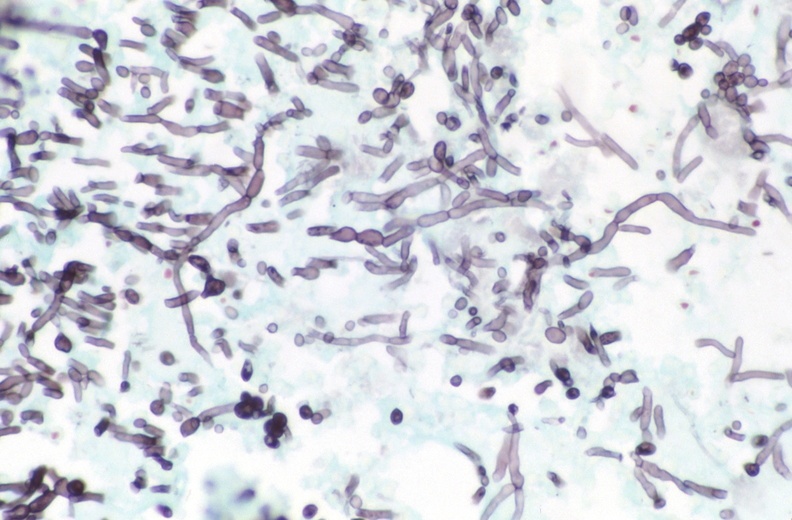}
  \caption{\textbf{Microscopy image for Pattern~6.} The image shows fungal hyphae and spores under microscopy -- no gastrointestinal tissue is present. \texttt{BioMedParseTextSeg} returned a large segmented area (121,312 pixels) when queried for ``gastrointestinal tissue,'' leading the model to a false positive conclusion.}
  \label{fig:case_pattern6}
\end{figure}

\noindent\textbf{Example.} The question ``Is gastrointestinal tissue present?'' (\cref{fig:case_pattern6}; ground truth: no) led \meissa{} to call \texttt{BioMedParseTextSeg} with query ``gastrointestinal tissue.'' The tool returned a large segmented area (121,312 pixels). Trusting this false positive, the model concluded: ``The area is significant, confirming its presence'' and answered ``Yes'' (incorrect).

In a similar case, the question ``Is free air present in this image?'' (ground truth: no) produced a segmented area of 5,192 pixels from BioMedParse. The model again trusted the tool output and answered ``Yes'' incorrectly.

\noindent\textbf{Analysis.} BioMedParse's text-prompted segmentation will produce non-zero area for many queries, regardless of whether the queried structure is actually present. The model lacks the calibration to distinguish genuine detections from false positives based on area magnitude or segmentation quality.

\paragraph{Pattern 7: ZoomIn Crash with Strategy Switch Recovery.}

\begin{figure}[ht]
  \centering
  \includegraphics[width=0.55\linewidth]{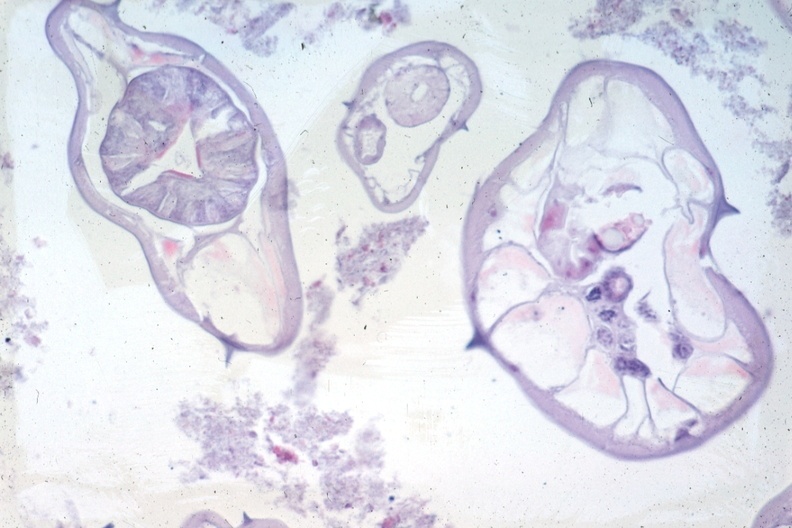}
  \caption{\textbf{Microscopy image for Pattern~7.} Histological cross-sections of pinworms (\textit{Enterobius vermicularis}). After 4 rounds of progressive zoom caused a crash, the model recovered by switching to \texttt{BioMedParseTextSeg}, correctly confirming pinworm presence.}
  \label{fig:case_pattern7}
\end{figure}

\noindent\textbf{Example.} The question ``Is pinworm present?'' (\cref{fig:case_pattern7}; ground truth: yes) began with 4 rounds of \texttt{ZoomInSubfigure}, progressively zooming into smaller regions until the crop became too small, triggering the error: ``ZoomIn failed: cannot write empty image as JPEG.'' The model recovered by switching strategy to \texttt{BioMedParseTextSeg} on the original image with query ``pinworm,'' which returned area=21,682. The model correctly answered ``Yes.''

\noindent\textbf{Analysis.} This demonstrates effective error recovery through strategy switching: when one tool fails due to parameter bounds, the model autonomously selects an alternative tool that operates on the original (uncropped) image. This recovery pattern is more common in the base model (which zooms more aggressively) than in \meissa{} (which has learned appropriate tool selection from training trajectories).

\subsection{MDAgents Trajectory Examples}
\label{sec:supp_mdagents_examples}

We present representative multi-agent collaboration trajectories from \meissa{}'s inference on PubMedQA, demonstrating the learned debate protocol.

\subsubsection{Example: Zero CLABSI Rate Sustainability (Correct, Unanimous)}

\noindent\textbf{Question:} ``Is zero central line-associated bloodstream infection rate sustainable?'' (Context: a NICU achieved zero CLABSI for $>$370 consecutive days, then $>$600 days after a brief uptick.)

\noindent\textbf{Ground truth:} Yes. \quad \textbf{Prediction:} Yes (correct).

\noindent\textbf{Difficulty Assessment} ($a_0$): The model classifies this as \texttt{intermediate}, reasoning: ``This question requires synthesis of statistical trends and operational policy management across multiple domains.''

\noindent\textbf{Expert Recruitment} ($a_1$): Four specialists are recruited: (1)~Pediatrician, (2)~Neonatologist, (3)~Infectious Disease Specialist, (4)~Critical Care Specialist.

\noindent\textbf{Expert Analysis} ($a_2$--$a_5$):
\begin{itemize}[leftmargin=*,itemsep=1pt]
\item \emph{Pediatrician}: ``Yes -- the study demonstrated sustained zero CLABSI for over 370 consecutive days after implementing evidence-based practice bundles and checklists.''
\item \emph{Neonatologist}: ``Yes -- despite a brief 3.3 per 1000 line-days increase in Q1 2013, the rate was successfully re-stabilized at zero after targeted interventions.''
\item \emph{Critical Care Specialist}: ``Yes -- a 77\% overall reduction over the 5-year study period demonstrates high sustainability.''
\end{itemize}

\noindent\textbf{Synthesis} ($a_6$): ``The implementation of evidence-based care bundles resulted in a sustained 600-day zero-rate, accompanied by a 77\% overall reduction. Despite a temporary increase requiring targeted corrections, the protocols maintained efficacy through interdisciplinary collaboration.'' Final answer: \textbf{Yes}.

\subsubsection{Example: Non-HDL-C as Cardiovascular Risk Predictor (Correct)}

\noindent\textbf{Question:} ``Is non-HDL-cholesterol a better predictor of long-term outcome after acute myocardial infarction compared to LDL-cholesterol?'' (Context: Thai patients post-AMI; non-HDL-C $>$130 mg/dL had HR 3.15 for MACE, while LDL-C $>$100 mg/dL paradoxically had HR 0.42.)

\noindent\textbf{Ground truth:} Yes. \quad \textbf{Prediction:} Yes (correct).

\noindent\textbf{Expert Recruitment}: Cardiologist, Cardiometabolic Specialist, Epidemiologist.

\noindent\textbf{Cross-Expert Debate}: All three experts independently concluded ``Yes.'' The Cardiometabolic Specialist highlighted the paradoxical LDL-C finding (HR 0.42) as a key reason to prefer non-HDL-C: ``Elevated LDL-C was unexpectedly associated with reduced MACE risk, undermining its utility as a standalone predictor.'' The Epidemiologist reinforced: ``Non-HDL-C showed a 3.15-fold increased risk, far exceeding LDL-C's predictive power.''

\noindent\textbf{Moderator}: Confirmed unanimous consensus.

\subsubsection{Example: Carotid Endarterectomy Risk in Octogenarians (Disagreement)}

\noindent\textbf{Question:} ``Are octogenarians at high risk for carotid endarterectomy?'' (Context: study of 103 procedures in 95 octogenarians with only 4 neurologic complications and 0 deaths.)

\noindent\textbf{Ground truth:} No. \quad \textbf{Prediction:} Yes (incorrect).

\noindent\textbf{Expert Analysis}: Cardiologist, Geriatrician, and Vascular Surgeon all concluded ``No,'' citing the 1.94\% complication rate as far below the commonly cited 15--25\% risk. However, during the debate phase, the Cardiologist briefly wavered, misinterpreting ``selected high-risk patients'' in the text as a diagnostic label rather than an enrollment criterion, before reverting to ``No.''

\noindent\textbf{Moderator Override}: Despite the 3-to-0 expert consensus of ``No,'' the Moderator concluded ``Yes,'' overriding the correct expert majority. This represents a failure mode where the synthesis agent fails to faithfully implement the specified majority-vote protocol.

\subsection{Clinical Simulation Trajectory Examples}
\label{sec:supp_agentclinic_examples}

We present representative clinical simulation trajectories from \meissa{}'s inference on MIMIC-IV, demonstrating the multi-turn diagnostic reasoning process.

\subsubsection{Example: Pulmonary Tuberculosis (Correct, 5 Tool Calls)}

\noindent\textbf{Patient Presentation:} 55-year-old male with chronic cough (8 weeks), productive of clear sputum, unintentional weight loss (10 lbs/3 months), night sweats, decreased appetite. Former smoker (20 pack-years, quit 5 years ago).

\noindent\textbf{Ground truth:} Pulmonary Tuberculosis. \quad \textbf{Prediction:} Tuberculosis (correct).

\begin{enumerate}[leftmargin=*,itemsep=2pt]
\item $a_0$: \texttt{RequestPhysicalExam}(\texttt{Vital\_Signs}) $\to$ $o_1$: Temperature 36.8$^\circ$C, BP 130/80, HR 82, RR 18.
\begin{quote}\small\textit{Reasoning:} ``Vital signs are within normal limits. No acute systemic infection or shock. The chronic cough, weight loss, and night sweats remain concerning for TB or lung cancer.''
\end{quote}

\item $a_1$: \texttt{RequestPhysicalExam}(\texttt{Respiratory\_Examination}) $\to$ $o_2$: Bilateral basal crackles (right-predominant), dullness at right base, increased tactile fremitus on right lower lobe.
\begin{quote}\small\textit{Reasoning:} ``These findings are highly suggestive of consolidation or mass in the right lower lobe, consistent with pneumonia, tuberculosis, or lung cancer.''
\end{quote}

\item $a_2$: \texttt{RequestTest}(\texttt{Chest X-Ray}) $\to$ $o_3$: Right upper lobe consolidation with cavitation, no pleural effusion.
\begin{quote}\small\textit{Reasoning:} ``Right upper lobe consolidation with cavitation -- a classic radiographic finding in TB. Cavitation is more commonly seen in TB than in malignancy, which typically presents with solid masses.''
\end{quote}

\item $a_3$: \texttt{RequestTest}(\texttt{Sputum Analysis}) $\to$ $o_4$: AFB smear positive, culture pending.
\begin{quote}\small\textit{Reasoning:} ``AFB smear positivity is diagnostic for TB in the context of clinical symptoms and radiographic findings. No further tests required.''
\end{quote}

\item $a_4$: \texttt{Terminate}(\texttt{diagnosis=``Tuberculosis''}).
\end{enumerate}

\noindent\textbf{Analysis.} The trajectory demonstrates systematic clinical reasoning: vitals $\to$ physical exam $\to$ imaging $\to$ confirmatory lab. A differential diagnosis (TB vs.\ lung cancer) is maintained throughout and resolved by the chest X-ray findings (cavitation favoring TB) and AFB smear confirmation.

\subsubsection{Example: Iron Deficiency Anemia (Correct, 3 Tool Calls)}

\noindent\textbf{Patient Presentation:} 55-year-old male with persistent fatigue (2 months), unintentional weight loss (10 lbs), occasional headaches, general weakness. Smoker (10 cigarettes/day). History of hypertension.

\noindent\textbf{Ground truth:} Iron Deficiency Anemia. \quad \textbf{Prediction:} Iron deficiency anemia (correct).

\begin{enumerate}[leftmargin=*,itemsep=2pt]
\item $a_0$: \texttt{RequestTest}(\texttt{Complete\_Blood\_Count}) $\to$ $o_1$: WBC 6,500, Hgb 10.2~g/dL (low), Hct 31\% (low), Plt 290K.
\begin{quote}\small\textit{Reasoning:} ``Microcytic anemia identified. Differential: iron deficiency, chronic disease, or hemolytic processes.''
\end{quote}

\item $a_1$: \texttt{RequestTest}(\texttt{Basic\_Metabolic\_Panel}) $\to$ $o_2$: Glucose 85, Na 138, K 4.2, Cr 1.0 (normal).
\begin{quote}\small\textit{Reasoning:} ``BMP normal, ruling out metabolic causes. Iron studies needed to confirm iron deficiency.''
\end{quote}

\item $a_2$: \texttt{RequestTest}(\texttt{Additional\_Tests}) $\to$ $o_3$: Serum iron 25~$\mu$g/dL (low), TIBC 400~$\mu$g/dL (high), ferritin 10~ng/mL (low).
\begin{quote}\small\textit{Reasoning:} ``Classic iron deficiency triad: low serum iron, elevated TIBC, low ferritin. Diagnostic criteria met.''
\end{quote}

\item $a_3$: \texttt{Terminate}(\texttt{diagnosis=``Iron deficiency anemia''}).
\end{enumerate}

\noindent\textbf{Analysis.} The model efficiently skips physical exam (appropriate for fatigue/weight-loss presentations) and proceeds directly to lab workup. Each test narrows the differential systematically: CBC confirms anemia, BMP excludes metabolic causes, iron studies confirm the specific etiology.

\subsubsection{Example: Acute Myocardial Infarction (Correct, 6 Tool Calls)}

\noindent\textbf{Patient Presentation:} 45-year-old male with sudden-onset central chest pain (1 hour), tight/squeezing, radiating to left arm, shortness of breath, nausea. No relief with rest. History of hypertension (10 years). Former smoker.

\noindent\textbf{Ground truth:} Acute Myocardial Infarction (STEMI -- Inferior Wall). \quad \textbf{Prediction:} Acute inferior myocardial infarction (correct).

\begin{enumerate}[leftmargin=*,itemsep=2pt]
\item $a_0$: \texttt{RequestPhysicalExam}(\texttt{Vital\_Signs}) $\to$ $o_1$: BP 145/90, HR 95, RR 20, Temp 36.8$^\circ$C.
\item $a_1$: \texttt{RequestPhysicalExam}(\texttt{Cardiac\_Examination}) $\to$ $o_2$: Normal S1/S2, no murmurs/rubs/gallops.
\item $a_2$: \texttt{RequestPhysicalExam}(\texttt{Respiratory\_Examination}) $\to$ $o_3$: Clear breath sounds bilaterally.
\item $a_3$: \texttt{RequestTest}(\texttt{Electrocardiogram}) $\to$ $o_4$: ST-segment elevation in leads II, III, aVF.
\item $a_4$: \texttt{RequestTest}(\texttt{Troponin I Levels}) $\to$ $o_5$: Elevated at 0.4~ng/mL (normal $<$0.01).
\item $a_5$: \texttt{Terminate}(\texttt{diagnosis=``Acute inferior myocardial infarction''}).
\end{enumerate}

\noindent\textbf{Analysis.} The model follows the standard chest-pain protocol: physical exams rule out valvular disease (no murmurs/gallops), pericarditis (no rubs), and pulmonary causes (clear lungs). ECG identifies the classic inferior STEMI pattern (ST elevation in II, III, aVF), confirmed by elevated troponin. This is the most thorough trajectory (6 tool calls), demonstrating the model's ability to utilize deep interaction depth ($T{=}6$) when clinically appropriate.

\bibliographystyle{splncs04}
\bibliography{main}
\end{document}